\documentclass[10pt,twocolumn,letterpaper]{article}

\usepackage{3dv}
\usepackage{times}
\usepackage{epsfig}
\usepackage{graphicx}
\usepackage{amsmath}
\usepackage{amssymb}

\usepackage[ruled]{algorithm2e} %
\usepackage{booktabs} %
\usepackage{graphicx}
\usepackage{makecell}
\usepackage{multirow}
\usepackage{wrapfig}

\usepackage{color}
\usepackage{steinmetz}
\usepackage{graphicx}
\usepackage{amssymb}
\usepackage{amsmath}
\usepackage{mathtools}
\usepackage{algorithmic}
\usepackage{newlfont} %
\usepackage{fixltx2e}
\usepackage[english]{babel} %
\usepackage{enumitem}
\usepackage{microtype}
\usepackage{floatflt}
\usepackage{booktabs} %
\usepackage{subcaption}
\usepackage{epsfig}
\usepackage{xcolor}
\usepackage{tabularx}
\usepackage{ragged2e}

\newcommand{\bff}{\mathbf{f}}

\newcommand{\bw}{\mathbf{w}}

\newcommand{\bC}{\mathbf{C}}

\newcommand{\bL}{\mathbf{L}}

\newcommand{\bT}{\mathbf{T}}

\newcommand{\mP}{\mathcal{P}}

\newcommand{\mM}{\mathcal{M}}
\newcommand{\mN}{\mathcal{N}}

\newcommand{\mS}{\mathcal{S}}

\newcommand{\mT}{\mathcal{T}}

\newcommand{\M}{\mathcal{M}}

\newcommand{\fig}[1]{Fig.~\ref{fig:#1}}
\newcommand{\mypara}[1]{\vspace{-4mm}\paragraph{#1}}

\usepackage[pagebackref=true,breaklinks=true,letterpaper=true,colorlinks,bookmarks=false]{hyperref}

\threedvfinalcopy %

\def\threedvPaperID{130} %

\ifthreedvfinal\fi
\begin{document}

\title{\vspace{-4mm}Learning Material-Aware Local Descriptors for 3D Shapes\vspace{-3mm}}

\author{Hubert Lin$^1$\\
\and
Melinos Averkiou$^2$ \\
\and
Evangelos Kalogerakis$^3$ \\
\and
Balazs Kovacs$^4$ \\
\and
Siddhant Ranade$^5$ \\
\and
Vladimir G. Kim$^6$ \\
\and
Siddhartha Chaudhuri$^{6,7}$ \\
\and
Kavita Bala$^1$ \\
\and
$^1$\small Cornell Univ.\\
\and
$^2$\small Univ. of Cyprus\\
\and
$^3$\small UMass Amherst\\
\and
$^4$\small Zoox\\
\and
$^5$\small Univ. of Utah\\
\and
$^6$\small Adobe\\
\and
$^7$\small IIT Bombay\\
}

\maketitle
\thispagestyle{empty}

\begin{abstract}

Material understanding is critical for design, geometric modeling, and analysis
of functional objects.
We enable material-aware 3D shape analysis by employing a projective
convolutional neural network architecture to learn material-aware descriptors
from view-based representations of 3D points for point-wise material
classification or material-aware retrieval.
Unfortunately, only a small fraction of shapes in 3D
repositories are labeled with physical materials, posing a challenge for
learning methods. To address this challenge, we crowdsource a dataset of $3080$
3D shapes with part-wise material labels. We focus on furniture models which
exhibit interesting structure and material variability. In addition, we also
contribute a high-quality expert-labeled benchmark of $115$ shapes from
Herman-Miller and IKEA for evaluation. 
We further apply a mesh-aware
conditional random field, which incorporates rotational and reflective
symmetries, to smooth our local material predictions across neighboring
surface patches. We demonstrate the effectiveness of our learned descriptors for
automatic texturing, material-aware 
retrieval, and physical simulation. The dataset and code will be publicly
available.

\end{abstract}

\begin{figure*}[t]
  \centering
  \includegraphics[width=0.9\linewidth]{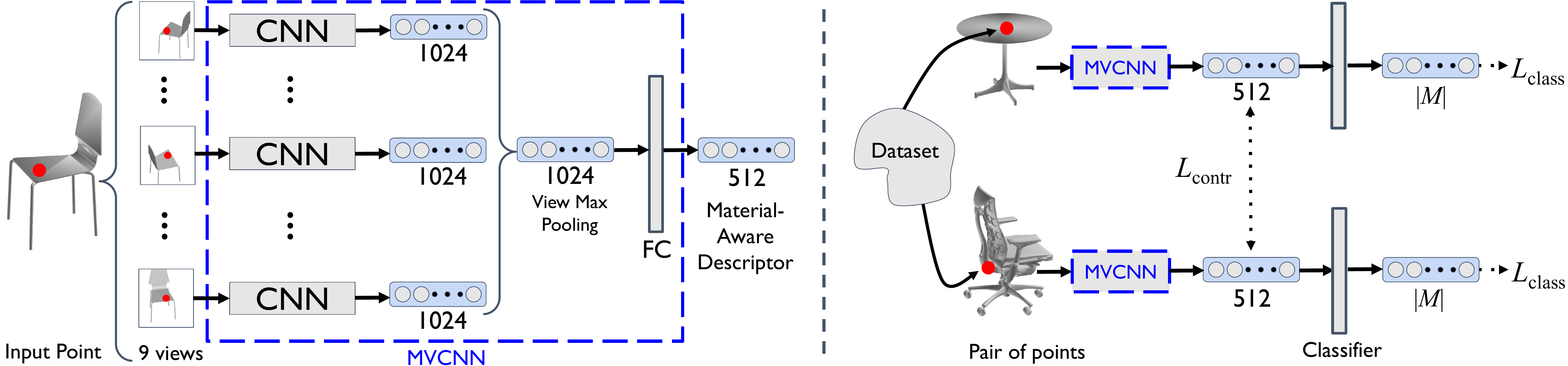}
  \caption{\small Left: Our
  MVCNN takes as input nine renderings of a point from multiple
views and scales, passes them through a CNN which produces 1024-dimensional
descriptors, which are then max-pooled over views and passed through a fully
connected layer which gives the final 512-dimensional descriptor. Right: The training
pipeline for pairs of points. The network is trained in a Siamese fashion. } \label{fig:arch}
\end{figure*}

\section{Introduction}

Materials and geometry are two essential attributes of objects that define their
function and appearance. The shape of a metal chair is quite different from that
of a wooden one for reasons of robustness, ergonomics, and manufacturability.
While recent work has studied the analysis and synthesis of 3D shapes
\cite{Mitra2013, yang2015reforming}, no prior work directly addresses the
inference of {\em physical} (as opposed to appearance-driven) materials from
geometry.

Jointly reasoning about materials and geometry can enable important
applications. Many large online repositories of 3D shapes have been developed
\cite{Shapenet2017}, but these lack tags that associate object parts with
physical materials which hampers natural queries based on materials, e.g.,
predicting which materials are commonly used for object parts,
retrieving objects composed of similar materials, and simulating how
objects behave under real-world physics.  Robotic perception often needs to
reason about materials: a glass tumbler must be
manipulated more gently than a steel one, and a hedge is a softer emergency
collision zone for a self-driving car than a brick wall. The color channel may
be unreliable (e.g., at night), and the primary input is geometric depth data
from LiDAR, time-of-flight, or structured light scanners. Interactive design
tools can suggest a feasible assignment of materials for fabricating a modeled
shape or indicate when a choice of materials would be unsuitable.

A key challenge in these applications is to reason about plausible
material assignments {\em from geometry alone}, since color textures are often
(in model repositories) or always (in night-vision robotics or interactive
design) absent or unreliable. Further, material choices are guided by
functional, aesthetic, and manufacturing considerations. This suggests that
material assignments cannot be inferred simply from physical simulations, but
require real-world knowledge.

In this paper, we address these challenges with a
novel method to compute {\em material-aware descriptors} of 3D points directly
from geometry. First, we crowdsource per-part
material labels for 3D shapes. Second, we train a projective convolutional
network ~\cite{Huang17MvCNN} to learn an embedding of geometric patches to a
material-aware descriptor space.  Third, we  curate a benchmark of shapes with
expert-labeled material annotations on which our material descriptors are
evaluated.

Learning surface point descriptors for 3D shape data has been explored in
previous approaches for 3D shape segmentation~\cite{Qi2017pointnet} and
correspondences~\cite{Zeng163DMatch,Huang17MvCNN}. However, there are challenges
with such an approach for our task. First, 3D shape datasets are limited in
size. While the largest image dataset with material labels comprises $\sim$437K
images~\cite{MINC15}, there is {\em no} shape dataset with material labels.
Second, many 3D shapes in repositories have missing, non-photorealistic, or
inaccurate textures (e.g., a single color for the whole shape). Material
metadata is rarely entered by designers for 3D models. Therefore, it is
difficult to automatically infer material labels. Third, gathering material
annotations is a laborious task, especially for workers who do not have a strong
association of untextured models with corresponding real-world objects.  We
address these challenges by designing a crowdsourcing task that enables
effective collection of material annotations.

\noindent Our contributions are the following:
\begin{itemize}
  \item \vspace{-2mm} The {\em first large database} of 3D shapes with per-part physical material labels, comprising a smaller expert-labeled benchmark set and a larger crowdsourced training set, and a {\em crowdsourcing strategy} for material labeling of 3D shapes.
\item \vspace{-3mm} A new deep learning approach for extracting {\em material-aware local descriptors} of surface points of untextured 3D shapes, along with a symmetry-aware CRF to make material predictions more coherent.
\item \vspace{-3mm} Prototype {\em material-aware applications} that use our descriptors for automatic texturing, part retrieval, and physical simulation.
\end{itemize}

\section{Previous work}
We review work on material prediction for shapes and images, as well as deep neural networks for shapes.

\mypara{Material prediction for shapes.} To make material assignment accessible
to inexperienced users Jain \etal~\cite{Jain12MaterialMemex} proposed a method
for automatically predicting reflective properties of objects. This method
relies on a database of 3D shapes with known reflective properties, and requires
data to be segmented into single-material parts. It uses light-field shape
descriptors~\cite{Chen03LightfieldDescriptors} to represent part geometry and a
graphical model to encode object or scene structure.  Wang
\etal~\cite{Wang16TexTransfer} proposed a method for transferring textures from
object images to 3D models. This approach relies on aligning a single 3D model
to a user-specified image and then uses inter-shape correspondence to assign the
texture to more shapes. Chen \etal~\cite{Chen15MagicDecorator} used color and
texture statistics in images of indoor scenes to texture a 3D scene.  They
require the scene to be segmented into semantic parts and labeled.  All these
methods focus on visual attributes and appeal of the final renderings. In our
work we focus on classifying physical materials, and do not assume any
additional user input such as an image that matches a 3D shape, or a semantic
segmentation.  Savva \etal~\cite{Savva2015semgeo} construct a large repository
of 3D shapes with a variety of annotations, including category-level priors over
material labels (e.g., ``chairs are on average 42\% fabric, 40\% wood, 12\%
leather, 4\% plastic and 2\% metal'') obtained from annotated image
datasets~\cite{Bell2013}. The priors are not specific to individual shapes. Chen
\etal~\cite{Chen2018text2shape} gather natural language descriptions for 3D
shapes that sometimes include material labels (``This is a brown wooden
chair''), but there is no fine-grained region labeling that can be used for
training.  Yang \etal~\cite{yang2015reforming} propose a data-driven algorithm
to reshape shape components to a target fabrication material. We aim to produce
component-independent material descriptors that can be used for a variety of
tasks such as classification, and we consider materials beyond wood and metal.

\mypara{Material prediction for images.}
Photographs are a rich source of the appearance of objects. Image-based material
acquisition and measurement has been an active area for decades; a comprehensive
study of image-based measurement techniques can be found in~\cite{WLLRZ09}.
Material prediction ``in the wild'', i.e., in uncontrolled non-laboratory
settings, has recently gained more interest fueled by the availability of
datasets like the Flickr Materials
Database~\cite{liu2010exploring,sharanIJCV2013}, Describable Textures
Dataset~\cite{dtd14}, OpenSurfaces~\cite{Bell2013}, and Materials in
Context~\cite{MINC15}. Past techniques identified features such as gradient
distributions along image edges \cite{liu2010exploring}, but recently deep
learning has set new records for material recognition (\cite{dtd14, MINC15}). In
our work, we focus on renderings of untextured shapes rather than photographs of
real world scenes.

\mypara{Deep neural networks for shape analysis.} A variety of neural network
architectures have been proposed for both global (classification) and local
(segmentation, correspondences) reasoning about 3D shapes. The variety of models
is in large part due to the fact that unlike 2D images, which are almost always
stored as raster grids of pixels, there is no single standard representation for
3D shapes. Hence, neural networks based on polygon
meshes~\cite{Masci2015,Boscaini2015spectral}, 2D
renderings~\cite{su15mvcnn,Kalogerakis17ShapePFCN,Huang17MvCNN}, local
descriptors after spectral alignment~\cite{Yi2016syncspeccnn}, unordered point
sets~\cite{wang2018dynamic, li2018pointcnn,
Qi2017pointnet,Qi2017pointnetpp,Su2017splatnet}, canonicalized
meshes~\cite{Maron17}, dense voxel
grids~\cite{Wu15VOLCNN,Girdhar16,Muralikrishnan2018}, voxel
octrees~\cite{Riegler17,Wang17}, and collections of surface
patches~\cite{Groueix2018}, have been developed. Bronstein
\etal~\cite{bronstein2017geometric} provide an excellent survey of spectral,
patch and graph-based approaches. Furthermore, methods such as
~\cite{litany2017deep} have been proposed for dense shape correspondences.  Our
goal is to learn features that reflect physical material composition, rather
than representations that reflect geometric or semantic similarity. Our specific
architecture derives from projective, or multi-view convolutional networks for
local shape analysis~\cite{Kalogerakis17ShapePFCN,Huang17MvCNN}, which are good
at identifying fine-resolution features (e.g., feature curves on shapes,
hard/smooth edges) that are useful to discriminate between material classes.
However, our approach is conceptually agnostic to the network used to process shapes, and
other volumetric, spectral, or graph-based approaches could be used instead.

\section{Data Collection}

\label{sec:data}

We collected a crowd-sourced training set of $3080$ 3D shapes annotated with
per-component material labels. We also created a  benchmark of $115$  3D
shapes with verified material annotations to serve as ground-truth.
Both datasets will be made publicly available.

\subsection{3D Training Shapes}
Our 3D training shapes originate from the ShapeNet v2
repository~\cite{Shapenet2017}. We picked shapes from three categories with
interesting material and structural variability: $6778$ chairs, $8436$ tables
and $1571$ cabinets. To crowd-source reliable material annotations for these
shapes, we further pruned the original shapes as follows.

First, observing that workers are error-prone on texture-less shapes, we removed
shapes that did not include any texture references. These account for $32.2\%$
of the original shapes.  Second, to avoid relying on crowd workers for tedious
manual material-based mesh segmentation,
we only included shapes with pre-existing components (i.e., groups in
their OBJ format).  We also removed over-segmented meshes ($>20$ components),
since these tended to have tiny parts that are too laborious to label. Meshes
without any, or with too many components accounted for an additional $17.1\%$ of
the original shapes.  Third, to remove low-quality meshes that often resulted in
rendering artifacts and further material ambiguity, we pruned shapes with fewer
than $500$ triangles/vertices (another $30.8\%$ of the dataset).
Finally, after removing duplicates, the remaining shapes were $1453$ chairs,
$1460$ tables, and $218$ cabinets, summing to a total of $3131$ shapes to be
labelled.
To gather material annotations for the components of these shapes, we created
questionnaires released through the Amazon Mechanical Turk (MTurk) service. Each
questionnaire had $20$ queries (see supplementary for interface).
Four different rendered views covering the front, sides and back of the textured
3D shape were shown. At the foot of the page, a single shape component was
highlighted.
Each query highlighted a different component. Workers were asked to select a
label from a set of materials $\M$ for the highlighted component. The set of
materials was $\M = \{${\em wood}, {\em plastic}, {\em metal},
{\em glass}, {\em fabric~(including leather)}$\}$. 
We selected this set to cover materials commonly found in furniture available in
ShapeNet. We deliberately did not allow workers to select multiple materials to
ensure they picked the most plausible material given the textured component
rendering. We also provided a ``null'' option, with associated text ``cannot
tell / none of the above'' so users could flag components whose
material they found impossible to guess. Our original list of materials also included
{\em stone}, but workers chose this option only for a small fraction 
of components ($\sim$0.5\%), and thus we excluded it from training and testing.
Preliminary versions of the
questionnaire showed that users often had trouble telling apart {\em metal} from
{\em plastic} components. 
We believe the reason was that {\em metal} and {\em plastic} components
sometimes have similar texture and color. Thus, we provided an
additional option ``{\em metal} or {\em plastic}''. We note that our training
procedure utilized this multi-material option, which is still partially
informative as it excludes other materials.

Out of the $20$ queries in each questionnaire, $3$ of them were ``sentinels''
i.e., components whose correct material was clearly visible, unambiguous and
confirmed by us. We used these sentinels to check worker
reliability.
Users who incorrectly labelled any sentinel or selected ``null''
for more than half of the questions were ignored. In total, $7370$
workers participated in our data collection out of which $23.7\%$ were deemed as
``unreliable''. All workers were compensated with $\$0.25$ for completing a
questionnaire.  On average, $5.2$ minutes were spent per questionnaire.
\begin{figure}[t!]
				\begin{center}
			 \includegraphics[width=0.49\columnwidth]{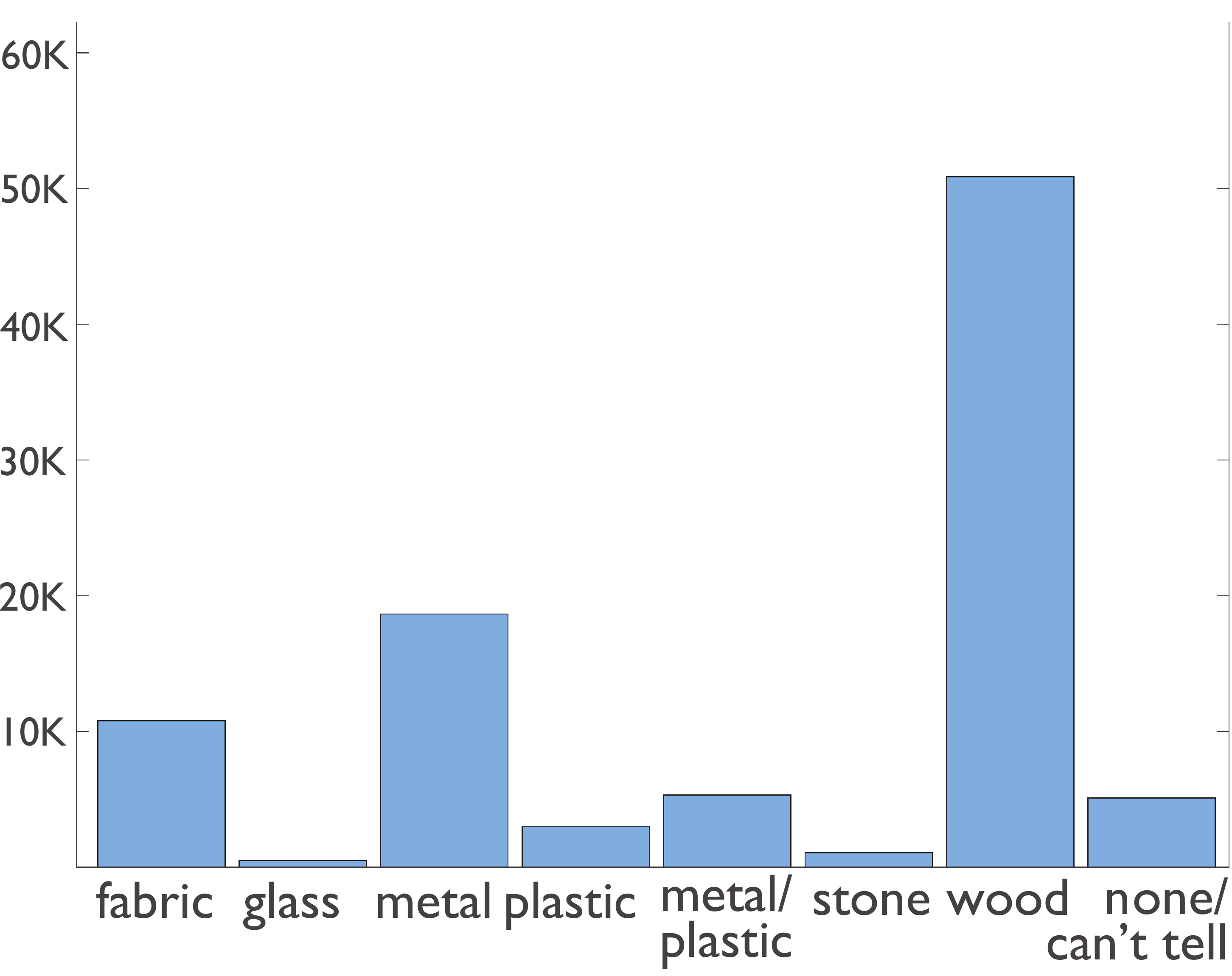}
			 \includegraphics[width=0.49\columnwidth]{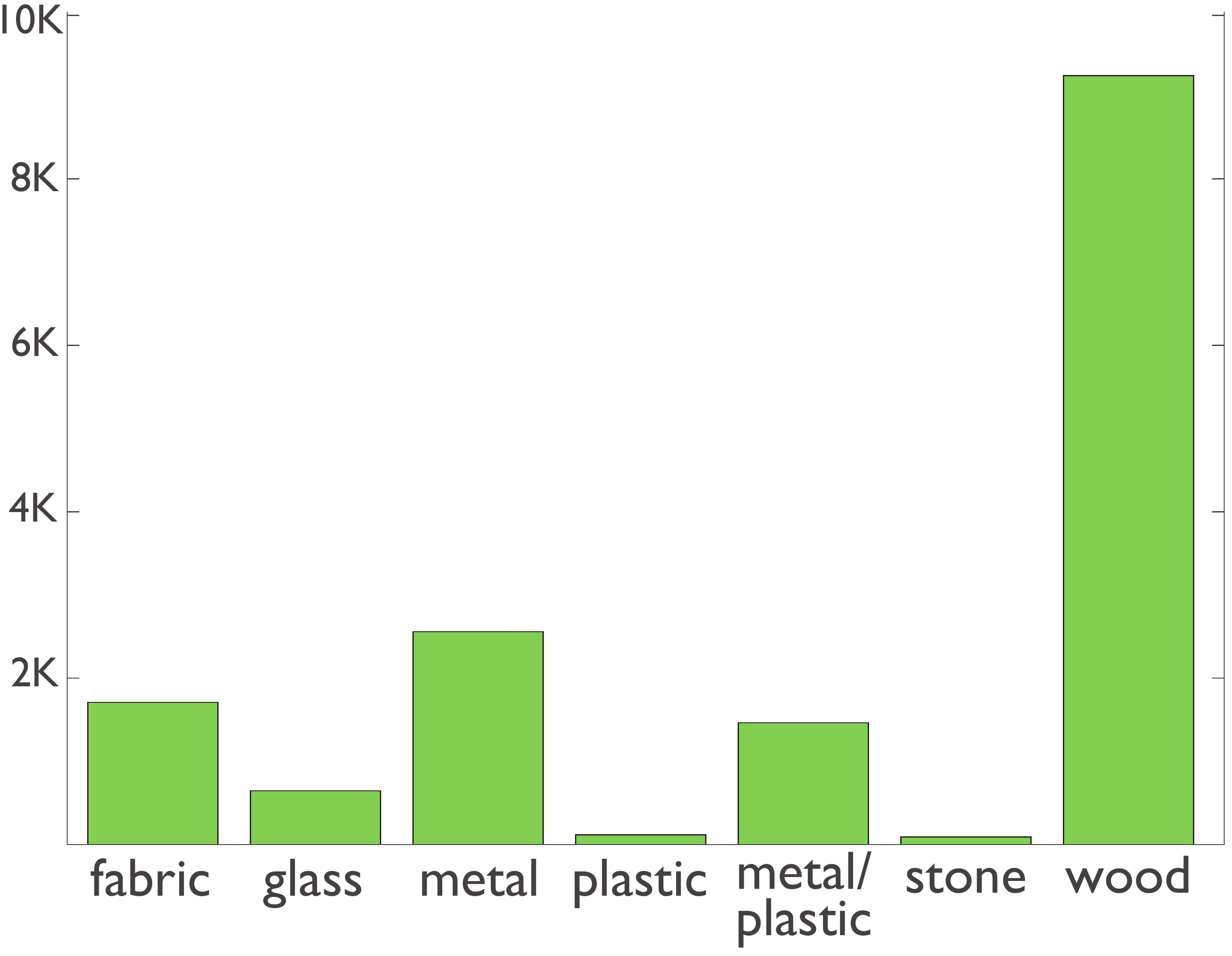}
				\vskip -1mm
				\caption{\small \label{fig:distribution3D} (a)~Distribution of answers in our questionnaires (b)~Distribution of material labels assigned to the components of our training dataset.}
				\end{center}
				\vskip -4mm
\end{figure}

Each component received $5$ answers (votes). 
The distribution of votes is shown in Figure ~\ref{fig:distribution3D}(a).
If $4$ or $5$ out of $5$ votes for a component agreed, we considered this
a consensus vote. $15835$ components received such consensus, of which
$547$ components had consensus on the ``null'' option.
Thus, $15288$ components (out of $19096$ i.e., $80.1\%$) acquired material
labels. We further checked and included $635$ components with transparent
textures and confirmed they were all glass. In total, we collected $15923$
labeled components in $3080$ shapes.
The distribution of material labels is shown in Figure~\ref{fig:distribution3D}(b).
For training, we kept only shapes with a majority of components labeled ($2134$ shapes).

\subsection{3D Benchmark Shapes}
\label{data-benchmark}

The 3D benchmark shapes originated from Herman Miller's online
catalog~\cite{HermanMiller} and 3D Warehouse~\cite{3DWarehouse}. All shapes were stored as
meshes and chosen because they had explicit references to product names and
descriptions from a corresponding manufacturer: IKEA~\cite{IKEA} or Herman Miller.
This dataset has $40$ chairs, $47$ tables and
$28$ cabinets. Expert annotators assigned material labels to all shape
components through direct visual reference to corresponding manufacturers'
product images as well as information from the textual product
descriptions. Such annotation is not scalable, hence this dataset is relatively
small and used purely for evaluation. See supplementary for distribution of
labeled parts.

\section{Network Architecture and Training}
%

Our method trains a convolutional network that embeds surface points of 3D
shapes in a high-dimensional descriptor space. To perform this embedding, our
network learns ``material-aware'' descriptors for points through a
multi-task optimization procedure.

\subsection{Network architecture}
To learn material-aware descriptors, we use the architecture visualized in
Figure \ref{fig:arch}.  The network follows a multi-view architecture
\cite{Kalogerakis17ShapePFCN,Huang17MvCNN}.
Other architectures could also be considered, e.g.
volumetric~\cite{Wu15VOLCNN,Zeng163DMatch},
spectral~\cite{Masci2015,Boscaini2015spectral,bronstein2017geometric}, or
point-based~\cite{Qi2017pointnet,Su2017splatnet}.

We follow Huang \etal's~\cite{Huang17MvCNN} multi-view architecture. We render
$9$ images around each surface point $s$ with a Phong shader and a single
directional light along the viewing axis. The rendered images depict local
surface neighborhoods around each point from distances of 0.25, 0.5 and 1.0
times the shape's bounding sphere radius.
The camera up vectors are aligned with the shape's upright
axis, as we assume shapes to be consistently upright-oriented. The viewpoints
are selected to maximize surface coverage and avoid self-occlusion
\cite{Huang17MvCNN}. In Huang \etal's architecture~\cite{Huang17MvCNN}, the
images per point are processed through AlexNet branches
\cite{Krizhevsky12ImageNet}
Because view-based representations for 3D
shapes are somewhat similar to 2D images, we chose to use GoogLeNet
\cite{szegedy2015going} instead, which achieved strong results for 2D material
recognition~\cite{MINC15}. Alternatives like VGG~\cite{Simonyan14vgg} yielded no notable
differences. We tried rendering $36$ views as in Huang \etal's
work, but since ShapeNet shapes are upright-oriented, we found that $9$
upright-oriented views were equivalent.

\begin{figure*}[ht!]
  \centering
  \includegraphics[width=0.8\linewidth]{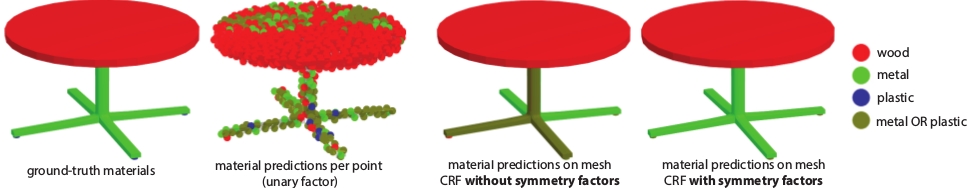}
  \vspace{-1mm}
  \caption{\small From left to right: Ground Truth, Per-Point Prediction from
  network, Post-CRF without symmetry factors, Post-CRF with symmetry factors.
Note that the material labeling of our CRF\ with symmetry is almost in total
agreement with ground-truth materials except for the plastic table leg caps
(bottom of the legs) which are plastic, but are labeled as ``metal or plastic'' from our CRF. }
  \label{fig:crf}
  \vspace{-2mm}
\end{figure*}

In our GoogLeNet-based MVCNN, we aggregate the 1024D output
from the 7x7 pooling layer after ``inception 5b'' for each of our $9$ views with
a max view-pooling layer \cite{su15mvcnn}.  This aggregated feature is reduced
to a 512D descriptor. A subsequent classification layer and sigmoid
layer compute classification scores. For training, all parameters are
initialized with the trained model from \cite{MINC15}, except for the
dimensionality reduction layer and classification layer whose parameters are
initialized randomly from a Gaussian distribution with mean $0$ and standard
deviation $0.01$.
\mypara{Structured material predictions.} Figure \ref{fig:crf} visualizes
the per-point material label predictions for a characteristic input mesh. Note
that self-occlusions and non-discriminative views can cause erroneous predictions.
Further, symmetric parts (e.g., left and right chair legs) lack consistency in material predictions, since
long-range dependencies between surface points are not explicitly considered in
our network. Finally, network material predictions are limited only to surface
points, and not throughout the whole shape.

To address these challenges, the last part of our architecture incorporates a
structured probabilistic model, namely a Conditional Random Field
\cite{Lafferty01} (CRF). The CRF models both local and long-range
dependencies in the material predictions across the input surface represented as
a polygon mesh, and also projects the point-based predictions onto the input
mesh. We treat the material predictions on the surface as binary random
variables. There are $|\mM|$ such variables per input polygon, each indicating the presence/absence of a particular material. Note that this formulation accommodates multi-material predictions.

Our CRF incorporates: (a) unary factors that evaluate the probability of
polygons to be labeled according to predicted point material labels,
(b) pairwise factors that promote the same material label for adjacent polygons
with low dihedral angle, (c) pairwise factors that promote the same material
label for polygons whose geodesic distance is small, (d) pairwise factors that
promote the same material label for polygons related under symmetry.
Specifically, given all surface random variables $\bC_s$ for an input shape
$s$, the joint distribution is expressed as follows:
\vspace{-1mm}
\begin{align*}
\small
P(\bC_s) \!=\! & \frac{1}{Z_s} \prod\limits_{m,f} \phi_{\textrm{unary}}(C_{m,f})
\prod\limits_{m,f,f' \in Adj} \phi_{\textrm{adj}}(C_{m,f}, C_{m,f'}) \\
\nonumber
& \nonumber
\prod\limits_{m,\textrm{}\,f,f'} \phi_{\textrm{dist}}(C_{m,f}, C_{m,f'})
\prod\limits_{m,\textrm{}\,f,f'} \phi_{\textrm{sym}}(C_{m,f}, C_{m,f'})
\vspace{-1mm}
\end{align*}
where $C_{m,f}$ is the binary variable indicating if face $f$ is
labeled with material $m$, and $Z_s$ is a normalization constant. The unary
factor sets the label probabilities of the surface point nearest to face $f$
according to the network output.
The pairwise factors $\phi_{\textrm{adj}}(C_{m,f},C_{m,f'})$ encode pairwise
interactions between adjacent faces, following previous CRFs for mesh
segmentation \cite{Kalogerakis17ShapePFCN}. Specifically, we define a factor
favoring the same material label prediction for neighboring polygons $(f,f')$
with similar normals. Given the angle
$\omega_{f,f'}$ between their normals ($\omega_{f,f'}$ is divided by $\pi$ to
map it between $[0,1]$), the factor is defined as follows:
\vspace{-1mm}
\begin{eqnarray*}
\small
\phi_{adj}(C_{m,f}\!=\!l,C_{m,f'}\!=\!l')\!=\!\\
&\nonumber\hspace{-30mm}\left\{
\begin{array}{l}
\!\!\!\!
\exp \!\Big(\!\!\!-\!w_{m,a} \!\cdot\! w_{m,l,l'} \!\cdot\! \omega_{f,f'}^2 \! \Big), \;\;\;\;\;\;\;\;\;l\!=\!l' \\
\!\!\!\!
\exp \!\Big(\!\!\! -\! w_{m,a} \!\cdot\! w_{m,l,l'}  \!\cdot\!  (1 \!-\! \omega_{f,f'} ^2)\!\Big),\;l \!\ne\! l'\\
\end{array} \right.
\vspace{-1mm}
\end{eqnarray*}
where $l$ and $l'$ represent the  $\{0,1\}$ binary labels for
adjacent faces $\{f,f'\}$, $w_{m,a}$ and $w_{m,l,l'}$ are learned factor- and
material-dependent weights.
The factors $\phi_{\textrm{adj}}(C_{m,f},C_{m,f'})$ favor similar labels for
polygons $f$, $f'$ which are spatially close (according to geodesic distance
$d_{f,f'}$) and also belong to the same connected component:
\vspace{-1mm}
\begin{eqnarray*}
\small
\phi_{dist}(C_{m,f}\!=\!l,C_{m,f'}\!=\!l')\!=\!\!\\
&\nonumber\hspace{-30mm}\left\{
\begin{array}{l}
\!\!\!\!
\exp \!\Big(\!\!\! -\! w_{m,d} \!\cdot\! w_{m,l,l'} \!\cdot\! d_{f,f'}^2 \! \Big),
\;\;\;\;\;\;\;\;\;\;l\!=\!l' \\
\!\!\!\!
\exp \!\Big(\!\!\! -\! w_{m,d} \!\cdot\! w_{m,l,l'}  \!\cdot\!  (1-d_{f,f'}^2 )\!\Big),\;l \!\ne\! l'
\end{array} \right.
\vspace{-1mm}
\end{eqnarray*}
where the weights $w_{m,d}$ and $w_{m,l,l'}$ are learned factor- and material-dependent parameters, and $d_{f,f'}$ represents the geodesic distance between $f$ and $f'$, normalized to $[0,1]$.

Finally, our CRF incorporates symmetry-aware factors. We note that such
symmetry-aware factors were not considered before in other CRF-based mesh
segmentation approaches.  Specifically, our factors
$\phi_{\textrm{symm}}(C_{m,f},C_{m,f'})$ favor similar labels for polygons $f$,
$f'$ which are related under a symmetry. We detect rotational and reflective
symmetries between components
 by
 matching surface patches through ICP, extracting their mapping transformations,
and grouping them together when they undergo a similar  transformation
following Lun \etal
\cite{Lun:2015:StyleSimilarity,Lun:2016:StyleTransfer}. The symmetry-aware factors are expressed
as:
\vspace{-1mm}
\begin{eqnarray*}
\small
\phi_{symm}(C_{m,f}\!=\!l,C_{m,f'}\!=\!l')\!=\!\! \\
&\nonumber\hspace{-30mm}\left\{
\begin{array}{l}
\!\!\!\!
\exp \!\Big(\!\!\! -\! w_{m,s} \!\cdot\! w_{m,l,l'} \!\cdot\! s_{f,f'}^2 \! \Big),
\;\;\;\;\;\;\;\;\;\;l\!=\!l' \\
\!\!\!\!
\exp \!\Big(\!\!\! -\! w_{m,s} \!\cdot\! w_{m,l,l'}  \!\cdot\!  (1-s_{f,f'}^2 )\!\Big),\;l \!\ne\! l'
\end{array} \right.
\vspace{-1mm}
\end{eqnarray*}
where the weights $w_{m,s}$ and $w_{m,l,l'}$ are learned factor- and label-dependent parameters, and $s_{f,f'}$ expresses the Euclidean distance between face centers after applying the detected symmetry.

Exact inference in this probabilistic model is intractable. Thus we use
mean-field inference to approximate the most likely joint assignment to all
random variables (Algorithm 11.7 of \cite{kf-pgmpt-09}). Figure
\ref{fig:crf} shows  material predictions  over the input mesh after performing
inference in the CRF with and without symmetry factors.

\subsection{Training}

To train the network, we sample $150$
evenly-distributed surface points from each of our 3D
training shapes. Points lacking a material label, or externally invisible,
are discarded.
The remaining points are subsampled to $75$ per shape
to fit memory constraints.
The network is trained end-to-end with a multi-task loss function
that includes a multi-class binary cross-entropy loss
for material classification and a contrastive loss \cite{hadsell2006dimensionality} to align 3D points in
descriptor space \cite{Masci2015} according to their underlying material (Figure \ref{fig:arch}
(right)). Specifically, given: (i) a set of training surface points $\mS$ from
3D shapes, (ii) a ``positive'' set $\mP$ consisting of surface point pairs
labeled with the same material label, (iii) a ``negative'' set $\mN$ consisting
of surface point pairs that do not share any material labels, (iv) binary
indicator values $t_{m,p}$ per training point $p \in \mS$ and label $m
\in \mM$ (equal to 1 when $p$ is labeled with label $m$, $0$
otherwise), the network parameters $\bw$ are trained according to the following
multi-task loss function:
\begin{equation*}
  \!\!\bL(\bw)\! =\! \lambda_{\text{class}} \bL_{\text{class}}(\bw) +
  \lambda_{\text{contr}}
  \bL_{\text{contr}}(\bw)
\end{equation*}

\noindent The loss function is composed of the following terms:
\begin{alignat*}{2}
  \!\!\bL_{\text{class}}(\bw)\! =&\! \sum\limits_{p \in \mS} \sum\limits_{m\in \mM}
  [t_{m,p} \log P(C_{m,p} =1 \mid \bff_p,\bw) + \\ \nonumber
  &(1-t_{m,p}) \log (1 - P(C_{m,p} =1 \mid\bff_p,\bw))]\\
  \!\!\bL_{\text{contr}}(\bw)\! &= [ \sum\limits_{p,q \in \mP} D^2(\bff_p,
    \bff_q) +\\ \nonumber
  &\hspace{10mm}\sum\limits_{p,q \in \mN} \max(M - D_{}(\bff_p, \bff_q),0)^2 ]
\end{alignat*}

where $P(C_{m,p} =1 \mid \bff_p,\bw)$ represents the probability of our network to
assign the material $m$ to the surface point $p$ according to its descriptor $\bff_p$,
$D^2(\bff_p, \bff_q)$\ measures squared Euclidean distances between the
normalized image and surface point descriptors, and $M$ is a margin typically used in
constrastive loss (we set it to $\sqrt{0.2} - 0.2$).
The loss terms have weights $\lambda_{\text{class}}=0.016~\&~
\lambda_{\text{contr}}=1.0$ which were selected empirically to balance the terms
to have same order of magnitude during training time.  We will refer to the
network optimized with this loss as ``Multitask''. We also experiment with a
variant that utilizes solely classification loss. In this case
$\lambda_{\text{class}}=1.0~\&~\lambda_{\text{contr}}=0.0$.  We will refer to
this network as ``Classification''.  Note that in both Multitask and
Classification, the classification layer is trained with an effective loss
weight of $\lambda_{\text{class}}=1.0$. For Multitask, the learning rate
multiplier of the classification layer is increased to compensate for
$\lambda_{\text{class}}=0.016$.

Multitask training is performed with Adam \cite{Adam14} with learning rate
$0.001$, $\beta_1=0.9,\beta_2=0.999$. The network is trained in cycles of $100K$
iterations. We choose a stopping point when losses converge
on a validation set, which occurs by the end of the second cycle. In order to
optimize for the contrastive loss, the network is trained in a Siamese fashion
with two branches that share weights (see Fig. \ref{fig:arch} right).
Classification training is performed through stochastic gradient descent with
momentum. The initial learning rate is set to $0.001$ and momentum is set to
$0.9$. The learning rate policy is polynomial decay with power $0.5$. $L_2$
weight decay is set to $0.0002$. We train Classification for two cycles of
$100K$ iterations. In the second cycle, the initial learning rate is reduced to
$0.0001$ and momentum is reduced to $0.4$. Note that variants of both
optimization procedures were tried for both loss functions and that we only
report the optimal settings here. We also note that we tried contrastive-only
loss but it did not perform as well as the variants here.

During training, point pairs are sampled from $\mP$ and $\mN$ with a
1:4 ratio with the intuition that learning to separate different
classes in descriptor space is more difficult than grouping the same
class together. To balance material classes, we explicitly cycle through all
material pair combinations when sampling pairs. For example, if we sample a
negative {\em wood-glass} pair, subsequent negative pairs will not be {\em
wood-glass} until all other combinations have been sampled. Because it is
possible for points to have multiple ground truth labels (e.g. {\em metal or
plastic}), we ensure that negative pairs do not share any ground truth labels.
For example, if we try to sample a {\em plastic-metal} pair and we draw a {\em
metal or plastic} point paired with a {\em metal} point, this pair would be
discarded and re-sampled until a true negative {\em plastic-metal} pair is
drawn. On a Pascal Titan X GPU, training with batchsize $5$ takes about $12$ hours per
$100K$ iterations.

\mypara{CRF training.} The CRF module is trained to maximize the log-likelihood
of the material labelings in our training meshes \cite{kf-pgmpt-09} on average:
\vspace{-2mm}
\begin{equation*}
L=\frac{1}{|\mT|}\sum\limits_{s \in \mT} \log P(\bC_s=\bT_s)
\vspace{-2mm}
\end{equation*}
where $\bT_s$ are ground-truth binary material labels per polygon in the
training shape $s$ from our training shape set $\mT$. We use gradient descent
and initialize the CRF weights to $1$~\cite{Kalogerakis17ShapePFCN}.
Training takes $\sim$8 hours on a Xeon E5-2630 v4 processor.

\section{Results} %
\subsection{Evaluation}
We evaluate our approach in a series of experiments.
For our test set, we sample $5$K evenly-distributed surface points from each of
$115$ benchmark test shapes. We discard externally invisible points, and evenly
subsample the rest to $1024$ points per shape. Our final test set consists of
$117$K points. See supplementary for distribution of points.

\mypara{Material-aware descriptors.} Mean precision for $k$ nearest neighbor
retrievals is computed by averaging the number of neighbors that share a ground
truth label with the query point over the number of retrieved points ($k$).
Nearest neighbors are retrieved in descriptor space from a class-balanced subset
of training points. Mean precision by class is computed by computing the mean
precision for subsets of the test set containing only test points that belong to
the class of interest. Table \ref{table:retrieval} summarizes the mean precision
at varying values of $k$. Both Classification and Multitask variations achieve
similar mean class precision at all values of $k$.  Furthermore, note that the
Multitask variation achieves better precision than the Classification variation
over all values of $k$ in every class except for wood. We believe this is likely
because the contrastive loss component of multi-task loss encourages distance
between clusters of dissimilar classes while classification-only loss encourages
the clusters to be separable without necessarily being far apart. Therefore it
is less likely for Multitask descriptors to have nearby neighbors from a
different cluster.

\begin{table}[t!]
  \small
  \noindent
  \begin{tabularx}{\linewidth}{|p{0.13\linewidth}||c|*{5}{>{\RaggedRight\arraybackslash}X|}}
   \hline
   & \makecell[c]{\!\!Mean\!\!} & \makecell[l]{\!\!Wood}&
   \makecell[l]{\!\!Glass} & \makecell[l]{\!\!Metal}&
   \makecell[l]{\!\!\!Fabric} & \makecell[l]{\!\!\!Plastic}\\
   \hline
   \hline
   \hspace{-0.5em} Classif.\!\! &&&&&& \\
   \hline
   \!\!\!\!$k$=1     & 55.7 & 76.4 & 34.3 & 65.0 & 56.1 & 46.7 \\
   \!\!\!\!$k$=30    & 56.9 & 75.3 & 41.1 & 64.9 & 55.3 & 47.6 \\
   \!\!\!\!$k$=100   & 57.3 & 75.1 & 43.0 & 64.9 & 55.5 & 48.0 \\
   \hline
   \hline
   \hspace{-5pt}Multitask\!\!&&&&&&\\
   \hline
   \!\!\!\!$k$=1   & 56.2 & 62.2 & 40.8 & 68.6 & 58.0 & 51.2 \\
   \!\!\!\!$k$=30  & 56.2 & 61.0 & 42.6 & 68.9 & 57.4 & 51.1 \\
   \!\!\!\!$k$=100 & 56.6 & 60.7 & 44.7 & 68.7 & 57.4 & 51.5 \\
   \hline
  \end{tabularx}
  \caption{\small Precision (\%) at $k$ nearest neighbors in descriptor space.
    \vspace{-2mm}}
  \label{table:retrieval}
\end{table}

\mypara{Material prediction.}
To demonstrate that our descriptors are useful for material classification,
we evaluate the learned classifier on our test shapes. We measure the top-1
accuracy per material label. The top-1 accuracy of a label
prediction for a given point is 1 if the point has that label according to
ground-truth, and 0 otherwise. If the point has multiple ground-truth labels,
the accuracy is averaged over them. The top-1 accuracy for a material
label is computed by averaging this measure over all surface points. These
numbers are summarized in Table \ref{table:acc}.

We note that both Classification and Multitask
variations produce similar mean class top-1 accuracies.  However, the Classification variation exhibits a
larger variance in its  top-1 class accuracies, with better wood accuracy in
exchange for worse glass and fabric accuracies compared to the Multitask variation.
After applying the CRF, both variations have improved top-1 accuracies for all classes
except for glass. Glass prediction accuracy remains almost the same for the Multitask variation, but drops
drastically for Classification. We suspect that this occurs because glass
parts  sometimes share   similar geometry  with wooden parts in furniture (for example,
flat tabletops or flat cabinet doors may be  made of either glass or wood). In this case, several point-wise
predictions will compete for both glass and wood.
If more of these predictions are wood rather than glass, it is likely the CRF will
smooth out the predictions towards wood, which will result in performance drop for glass predictions.  \fig{cm} shows top-1
prediction confusion matrices. Wood points are often predicted correctly, yet sometimes are confused with  metal. Glass points are often confused with wood points.   Fabric is occasionally confused with
plastic or wood. These confusions often happen for chairs with thin leather backs or
thin seat cushions. Plastic is occasionally confused with metal. This is due to
parts that are thin rounded cylinders often  used  in both metal and plastic-made components.  Furthermore, the proportion of {\em
plastic} labels to {\em ``metal or plastic''} labels is low in our training dataset, which makes the learning less reliable in the case of plastic.  In both variations, there is a bias towards wood predictions. This is
likely due to the abundance of wooden parts in real-world furniture designs
reflected in our datasets. However, the bias is less pronounced in the Multitask variation. Thus we believe that the Multitask variation is better for a more  balanced generalization performance across  classes.

\begin{table}[t!]
 \small
\begin{tabularx}{\linewidth}{|p{0.15\linewidth}||c|*{5}{>{\RaggedRight\arraybackslash}X|}}
   \hline
   \makecell[c]{\!\!Network\hspace{-1em}} & \makecell[c]{\!\!Mean\!\!} & \makecell[l]{\!\!Wood}&
   \makecell[l]{\!\!Glass} & \makecell[l]{\!\!Metal}&
   \makecell[l]{\!\!\!Fabric} & \makecell[l]{\!\!\!Plastic}\\
   \hline
   \hline
   \makecell[l]{\!\!MINC-bw} & 38 & 1.2 & 38 & 65 & 20 & 65 \\
   \hline
   \makecell[l]{\!\!Classif.} &  65 & 82 & 53 & 72 & 62 & 55 \\
   \makecell[l]{\!\!C+CRF} & 66 & 85 & 36 & 77 & 66  & 65 \\
   \hline
   \makecell[l]{\!\!\!Multitask} &  66 & 68 & 65 & 72 & 70 & 53 \\
   \makecell[l]{\!\!\!MT+CRF}  & 71 & 75 & 64 & 74 & 74 & 68 \\
   \hline
\end{tabularx}
\caption{\small Top-1 material classification accuracy (\%). Baseline MINC-bw is
        MINC~\cite{MINC15} trained on greyscale photos and tested on untextured 3D
        renderings.\vspace{-2mm}}
\label{table:acc}
\end{table}

\begin{figure}[t!]
  \centering
  \begin{subfigure}[b]{0.49\linewidth}
      \includegraphics[width=\linewidth]{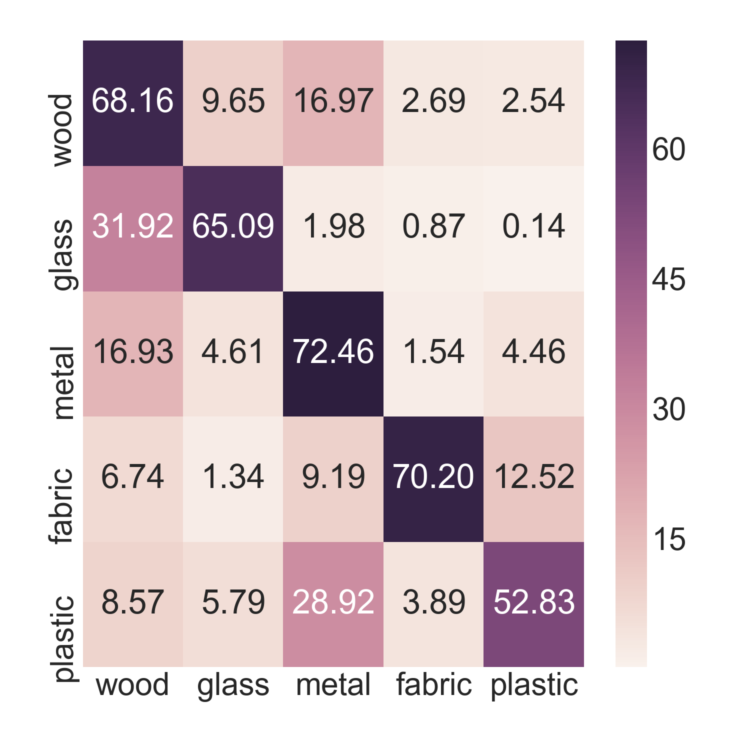}
      \caption{\small Multitask}
  \end{subfigure}
  \begin{subfigure}[b]{0.49\linewidth}
      \includegraphics[width=\linewidth]{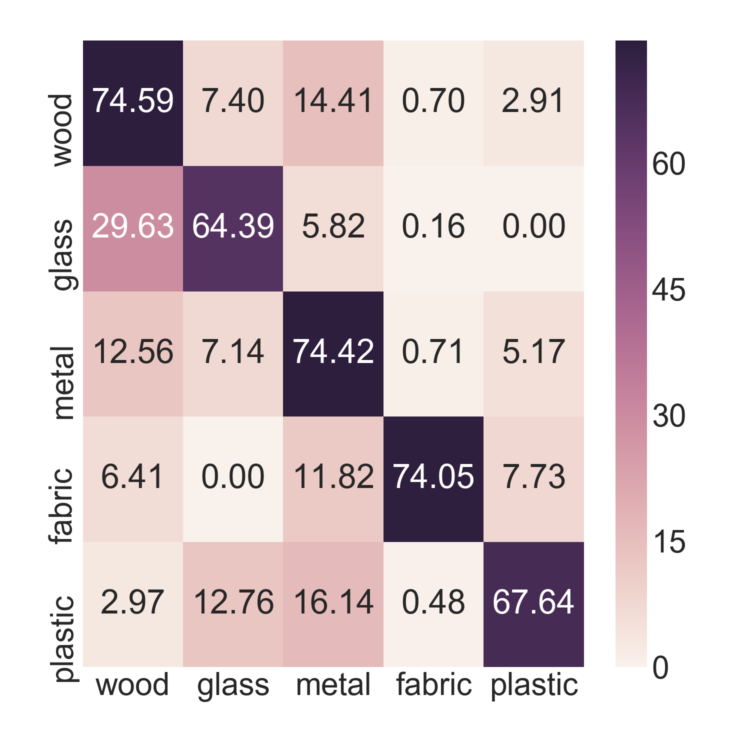}
      \caption{\small Multitask + CRF}
  \end{subfigure}
  \caption{\small Confusion matrices for Top-1 classification predictions.
    Rows are ground truths and columns are predictions.}
  \label{fig:cm}
\end{figure}

\begin{table}[b!]
\small
\begin{tabularx}{\linewidth}{|p{0.14\linewidth}||c|*{5}{>{\RaggedRight\arraybackslash}X|}}
   \hline
   \makecell[c]{\!\!Network\hspace{-1em}} & \makecell[c]{\!\!Mean\!\!} & \makecell[l]{\!\!Wood}&
   \makecell[l]{\!\!Glass} & \makecell[l]{\!\!Metal}&
   \makecell[l]{\!\!\!Fabric} & \makecell[l]{\!\!\!Plastic}\\
   \hline
   \hline
   \makecell[l]{\!\!C 3view} &  59 & 81 & 41 & 71 & 60 & 40 \\
   \makecell[l]{\!\!C 9view} &  65 & 82 & 53 & 72 & 62 & 55 \\
   \hline
   \makecell[l]{\!\!\!MT 3view} &  56 & 45 & 71 & 85 & 65 & 15 \\
   \makecell[l]{\!\!\!MT 9view} &  66 & 68 & 65 & 72 & 70 & 53 \\
   \hline
\end{tabularx}
\caption{\small \vspace{-3mm} Top-1 classification accuracy (\%) for 3 views vs 9
  views.}
\label{table:num_views}
\end{table}

\mypara{Effect of Number of Views.}
To study the effect of the number of views, we train the MVCNN with 3 views (1
viewpoint, 3 distances) and compare to our results above with 9 views (3
viewpoints, 3 distances): see Table
~\ref{table:num_views}. Multiple viewpoints are advantageous.

%
\subsection{Material-aware Applications}
\noindent{We illustrate the utility of the material-aware descriptors learned by our method in some prototype applications.}

\begin{figure}[t!]
  \centering
  \begin{subfigure}[b]{0.45\linewidth}
      \includegraphics[width=\linewidth]{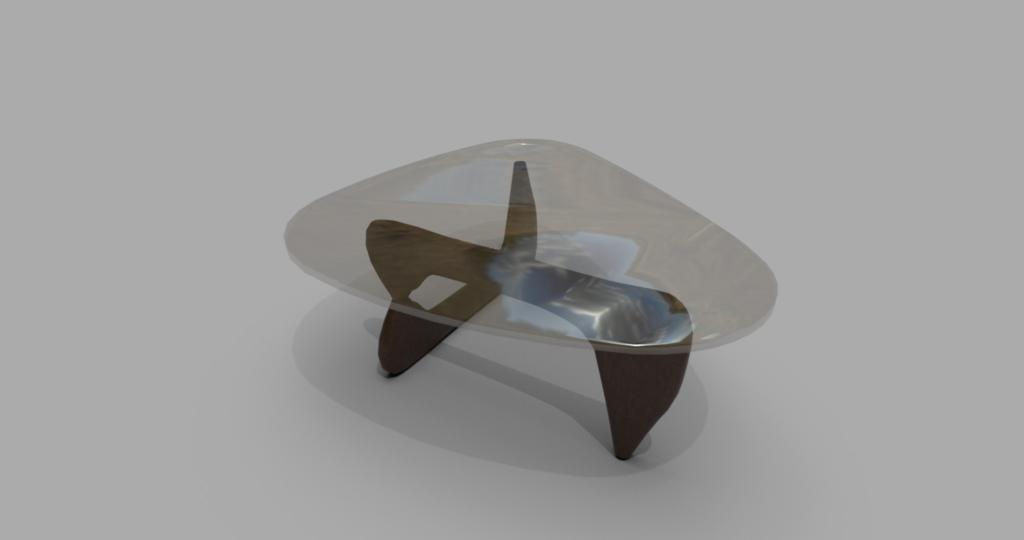}
  \end{subfigure}
  \begin{subfigure}[b]{0.45\linewidth}
      \includegraphics[width=\linewidth]{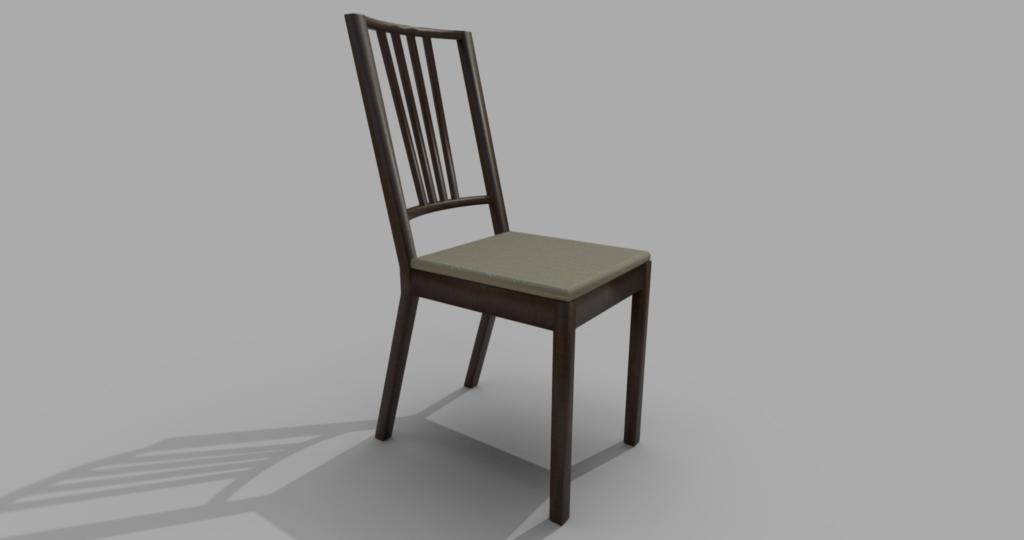}
  \end{subfigure}
  \begin{subfigure}[b]{0.45\linewidth}
     \includegraphics[width=\linewidth]{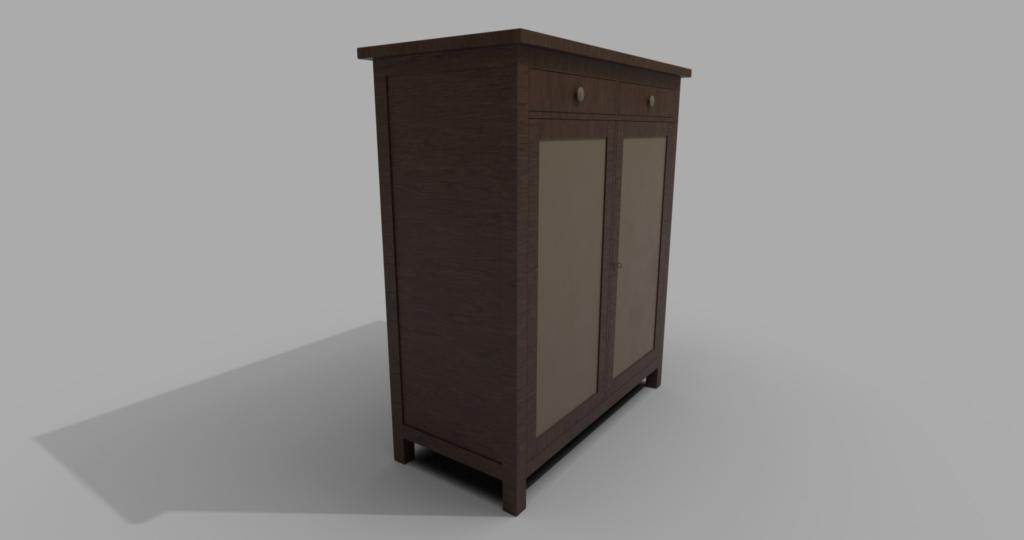}
  \end{subfigure}
  \begin{subfigure}[b]{0.45\linewidth}
     \includegraphics[width=\linewidth]{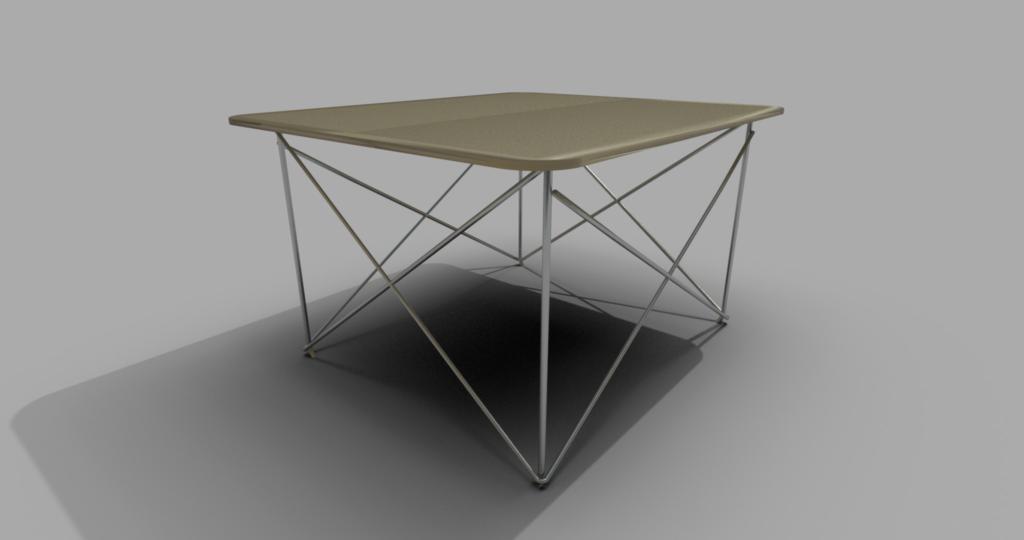}
  \end{subfigure}
  \caption{\small After material-aware segmentation of these representative shapes from our test set, shaders and textures corresponding to {\em wood}, {\em glass}, {\em fabric}, {\em plastic} and {\em metal} were applied to their faces.}
  \label{fig:autotexture}
\end{figure}

\mypara{Texturing.}
Given the material-aware segmentation produced by our method, we can automatically texture a 3D mesh based on the predicted material of its faces.
Such a tool can be used to automate texturing for new shapes or for collectively texturing existing shape collections.
If the mesh does not have UV coordinates, we generate them automatically by simultaneous multi-plane unwrapping. Then, we apply a texture to each mesh face according to the physical material predicted by the material-aware segmentation. We have designed representative textures for each of the physical materials predicted by our method (wood, plastic, metal, glass, fabric). Resulting renderings for a few of the meshes from our test set can be seen in Figure \ref{fig:autotexture}.

\mypara{Retrieval of 3D parts.}
\begin{figure}[t!]
  \centering
  \includegraphics[width=0.8\columnwidth]{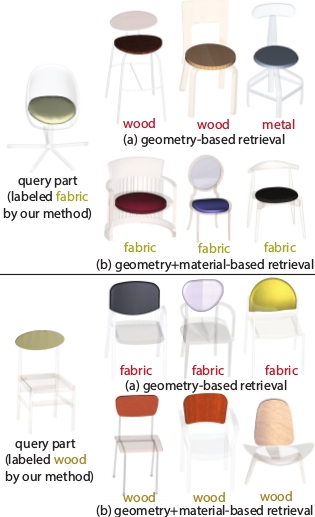}
  \caption{ \small Given a query (untextured) part of a 3D shape and its material label predicted by our method, we can search for geometrically similar parts by considering (a)  geometric descriptors alone, or (b)  geometric descriptors together with material labels. Note that we discard any retrieved parts that are near duplicates of the query parts to promote more diversity in the retrieved results.}
\label{fig:material_retrieval}
\vspace{1mm}
\end{figure}
Given a query 3D shape part from our test set, we can search for geometrically
similar 3D parts in the training dataset.
However, retrieval based on a geometric descriptor can return
parts with inconsistent materials (see Figure \ref{fig:material_retrieval}(a))
whereas a designer
might want to find geometrically similar parts with consistent materials
(e.g. to replace the query part or its texture
with a compatible database part)
In Figure \ref{fig:material_retrieval}(b), we show retrieval
results when we use both a geometric descriptor along with a simple material
compatibility check. Our pipeline
is used to obtain the material label for the untextured query part.
Then, we retrieve geometrically similar parts from our training set whose
crowdsourced material label agrees with the predicted one. For our prototype, we
used the multi-view CNN of Su \etal~\cite{su15mvcnn} to compute geometric
descriptors of parts.

\begin{figure}[t!]
  \centering
  \begin{subfigure}[b]{0.49\linewidth}
    \centering
    \includegraphics[width=0.9\linewidth]{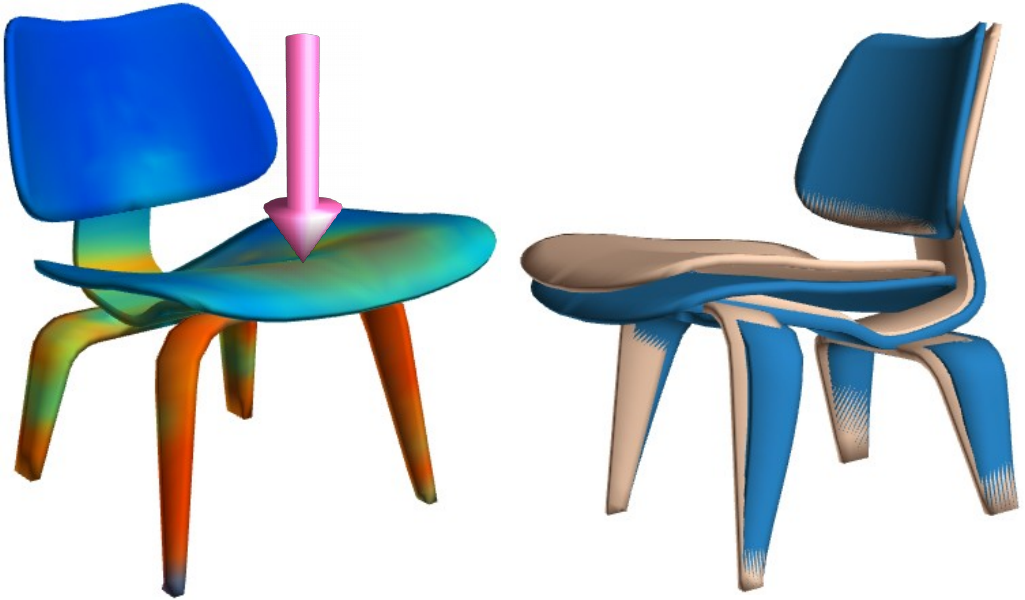}
    \caption{ \small Wood: \mbox{$\rho = 900$ kg/m$^3$,~~~~~~~} \mbox{$E = 12.6$ GPa}, \mbox{$\nu = 0.45$}}
  \end{subfigure}
  \begin{subfigure}[b]{0.49\linewidth}
    \centering
    \includegraphics[width=0.9\linewidth]{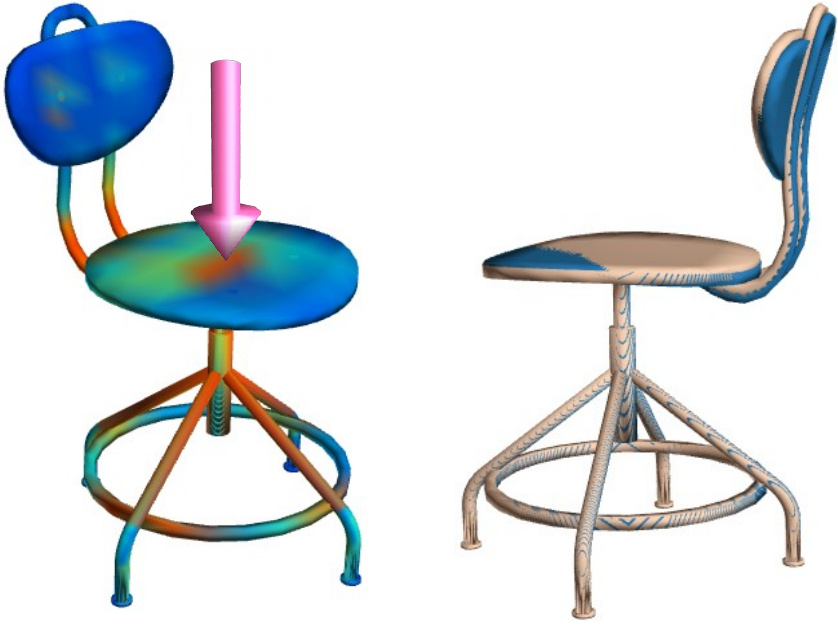}
    \caption{ \small Metal: \mbox{$\rho = 8050$ kg/m$^3$,~~~~~~~} \mbox{$E = 200$ GPa}, \mbox{$\nu = 0.3$}}
  \end{subfigure}
  \caption{\small A downward impulse of 4800 N$\cdot$s distributed over a chair seat (pink arrow) induces deformation (ignoring fracture and cavities). The left images show surface strain (area distortion), with blue = low and red = high. The corresponding deformation is visualized in the right images, with beige indicating the original shape and blue the overlaid deformed result. Wood, with much lower density ($\rho$) and Young's modulus ($E$), and higher Poisson's ratio ($\nu$), is more strongly deformed than the stiffer metal (steel). }
  \label{fig:simulation}
  \vspace{1mm}
\end{figure}

\mypara{Physical Simulation.}
Our material prediction pipeline allows us
to perform simulation-based analysis of raw geometric shapes without any manual
annotation of density, elasticity or other physical properties. This kind of
visualization can be useful in interactive design applications to assist
designers as they create models.
In Figure \ref{fig:simulation}, we show a prototype application which takes as
input an unannotated polygon mesh, and simulates the
effect of a downward force on it assuming the ground contact points are fixed.
The material properties of
the shape are predicted using a lookup table
which maps material labels predicted by our method to representative density and elasticity
values. We use the Vega toolkit~\cite{Vega} to select the force application
region and deform the mesh under a downward impulse of 4800 N$\cdot$s evenly
distributed over this area.
For this prototype, we
ignore fracture effects and internal cavities, and assume the material is
perfectly elastic. An implicit Newmark integrator performs finite element
analysis over a voxelized ($100^3$) version of the shape. The renderings in
Figure \ref{fig:simulation} show both the local surface strain (area
distortion) as well as the induced deformation of shapes with different
material compositions.

\section{Conclusion}

We presented a supervised learning pipeline to compute material-aware local descriptors for untextured 3D shapes, and developed the first crowdsourced dataset of 3D shapes with per-part physical material labels. Our learning method employs a projective convolution network in a Siamese setup, and material predictions inferred from this pipeline are smoothed by a symmetry-aware conditional random field. Our dataset uses a carefully designed crowdsourcing strategy to gather reasonably reliable labels for thousands of shapes, and an expert labeling procedure to generate ground truth labels for a smaller benchmark set used for evaluation. We demonstrated prototype applications leveraging the learned descriptors, and are placing the dataset in the public domain to drive future research in material-aware geometry processing.

Our work is a first step and has several limitations. Our experiments have studied only a small set of materials, with tolerably discriminative geometric differences between their typical parts. Our projective architecture depends on rendered images and can hence process only visible parts of shapes. Also, our CRF-based smoothing is only a weak regularizer and cannot correct gross inaccuracies in the unary predictions. Addressing these limitations would be promising avenues for future work.

We believe that the joint analysis of physical materials and object geometry is an exciting and little-explored direction for shape analysis and design. Recent work on functional shape analysis~\cite{Hu2018} has been driven by priors based on physical simulation, mechanical compatibility or human interaction. Material-aware analysis presents a rich orthogonal direction that directly influences the function and fabricability of shapes. It would be interesting to combine annotations from 2D and 3D datasets to learn better material prediction models. It would also be interesting to reason about parametrized or fine-grained materials, such as different types of wood or metal, with varying physical properties. As a driving application, interactive modeling tools that provide continuous material-aware feedback on the shape being modeled could significantly aid real-world design tasks. Finally, there is significant scope for developing ``material palettes'' -- probabilistic models of material co-use that take into account many intersectional design factors such as function, aesthetics, manufacturability and cost.

\paragraph{Acknowledgements.}
We acknowledge support from NSF (CHS-1617861, CHS-1422441, CHS-1617333). We thank Olga Vesselova for her input to the user study and Sourav Bose for help with FEM simulation of furniture.

\paragraph{Supplementary Material}
\section{Supplementary Material}

This supplementary material is organized as follows. First we show the data
collection interface and discuss additional statistics that may be of interest
(section ~\ref{sec:data}).  Second, we discuss some additional training details
(section ~\ref{sec:training}). Third, we show statistics for our test set
(section ~\ref{sec:testset}).
Fourth, we discuss in detail the 2D classification
baseline that we used in our evaluation (section ~\ref{sec:baseline}). Fifth,
we visualize embedding plots via t-SNE for our learned descriptor space (section
~\ref{sec:tsne}). Sixth, we show confusion matrices for both Classification and
Multitask networks (section ~\ref{sec:cm}), as well as for 3-view variants
(section ~\ref{sec:cm-3view}), and for network trained with only contrastive
loss (section ~\ref{sec:cm-contrastive}). Seventh, we show a sample of our dataset as
well as a visual sample of our material prediction results (section
~\ref{sec:data-vis}).

\subsection{\label{sec:data}Data collection}

\begin{figure}[ht!]
        \begin{center}
      \includegraphics[width=\columnwidth]{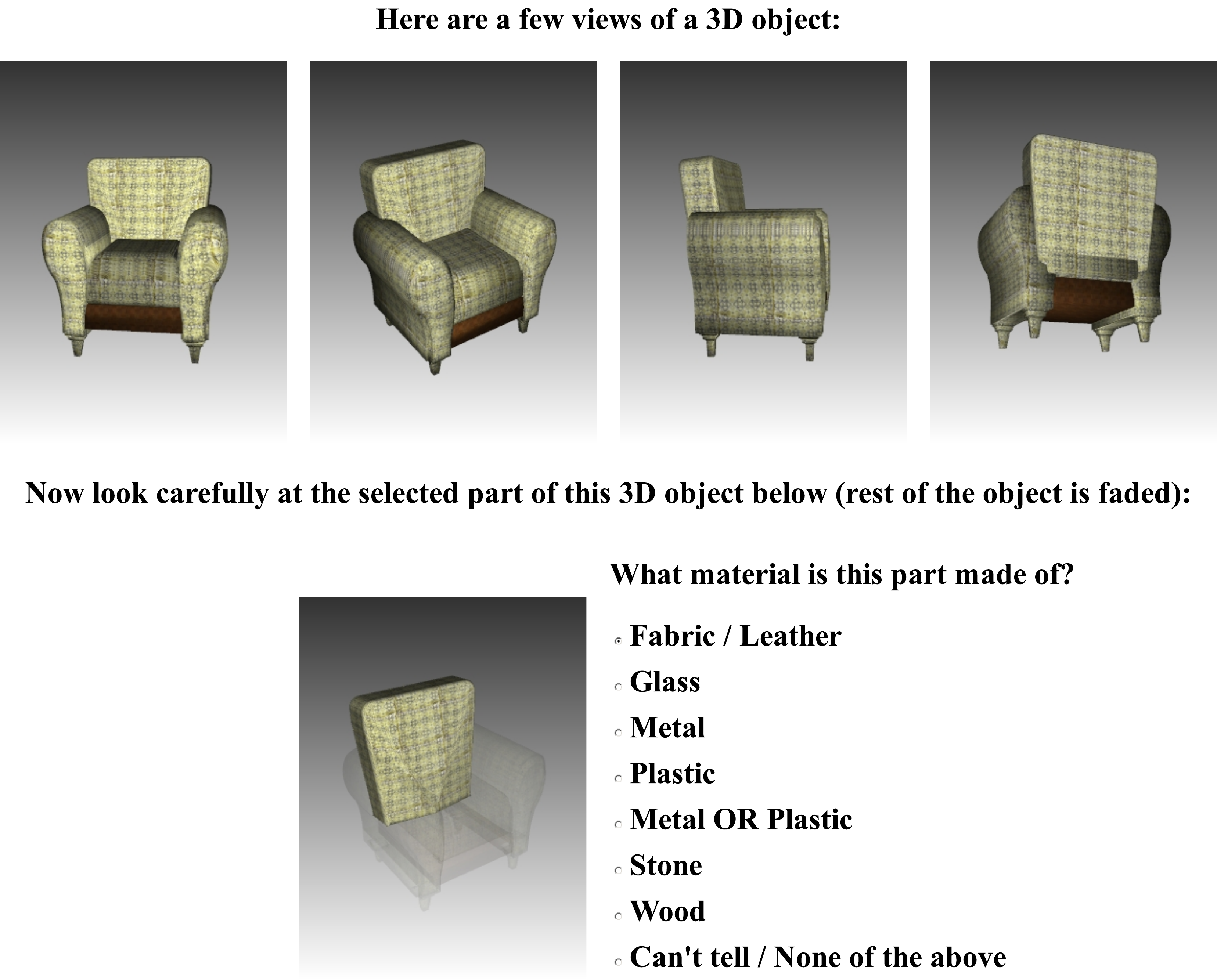}
        \vskip -1mm
        \caption{\label{fig:gallery} Our interface for collecting material labels for 3D\ shapes.}
        \end{center}
        \vskip -4mm
\end{figure}

Our data collection interface is shown in Fig.~\ref{fig:gallery}. Four different
rendered views covering the front, sides and back of the textured 3D shape were
shown. At the foot of the page, a single shape component was highlighted while
the rest of the 3D shape appeared faded.  Each query highlighted a different
component. Workers were asked to select a label from a set of materials for the
highlighted component. 
In total, we collected $15923$ labeled components in
$3080$ shapes. On average  $76\%$ of the surface area per mesh was labeled. For
training, we kept only shapes with $>50\%$ of components labeled ($2134$
shapes).

\subsection{\label{sec:training}Training Points}

To train the network, we sample $150$ evenly-distributed surface points from
each of our 3D training shapes. Points lacking a material label, or externally
invisible, are discarded. Point visibility is determined via ray-mesh
intersection tests. The remaining points are subsampled to $75$ per shape. This
subsampling is again performed so that selected points are approximately
uniformly distributed along the shape surface. The choice to sample $75$ points
per shape is due to memory limitations (we store the dataset in the main memory
to avoid slow I/O during training). The views corresponding to these points are
preprocessed and saved as single channel, unsigned integer arrays which are read
directly into memory at training time to prevent I/O bottlenecks.  Note that
sampling roughly 75 points per shape requires $\sim$ 60G memory. Preprocessing to store
into memory rather than reading from disk offered us a $~5-10$x speedup in
training time.



\subsection{\label{sec:testset}Benchmark Test Set Distributions}

\begin{figure}[th!]
				\begin{center}
			 \includegraphics[width=0.99\columnwidth]{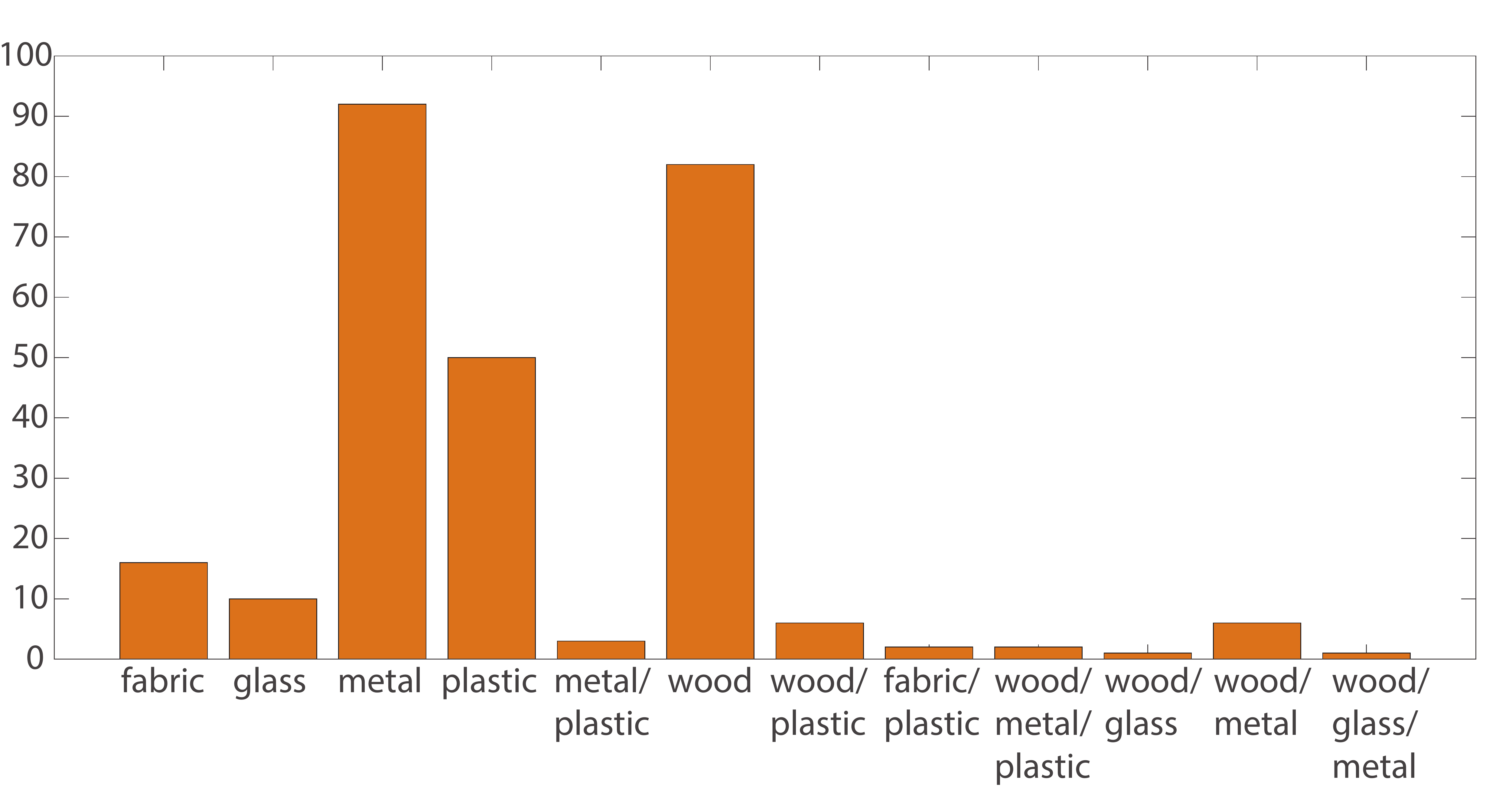}
			 \includegraphics[width=0.99\columnwidth]{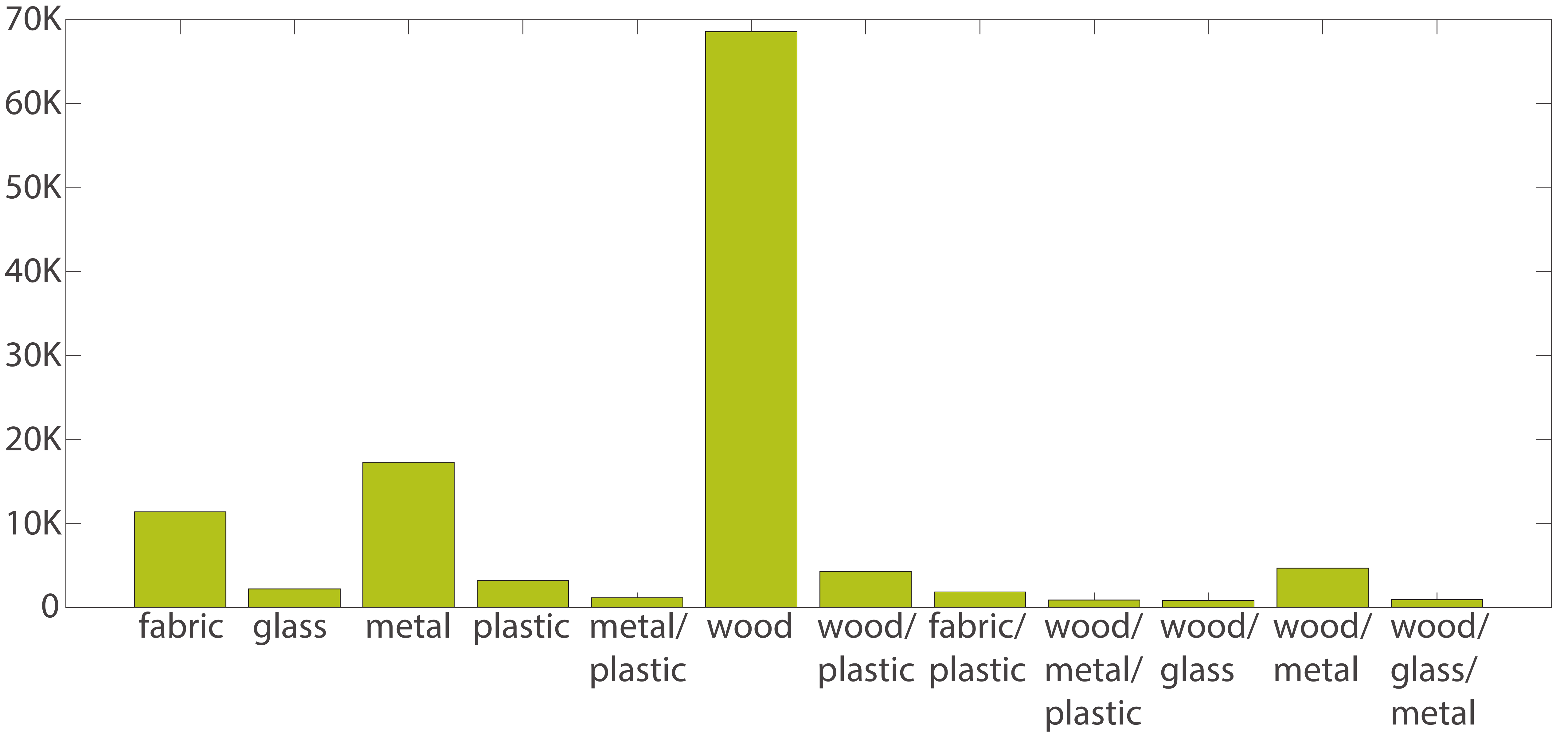}
				\vskip -1mm
				\caption{\small \label{fig:distribution3D_testset} (top)~Distribution of
			material labels assigned to components in our expert-labeled benchmark
		test shapes (bottom)~Distribution of material labels assigned to the points sampled
	from the benchmark test shapes that form our test set.}
				\end{center}
				\vskip -4mm
\end{figure}

In Fig.~\ref{fig:distribution3D_testset}, we show the distribution of labels
across components in the benchmark shape dataset, as well as the distribution of
labels across the points sampled from these shapes that form our evaluation test
set. Notice that although there are a large number of metal and plastic
components, relatively few metal or plastic points are sampled. This is because
many metal or components are small thin structures (e.g. handles, table legs).
Recall that we sample our test points uniformly across the surface of our shapes
and thus the surface area of a component is proportional to the number of points
sampled from that component.

\subsection{\label{sec:baseline}2D Classification Network Baseline}

\begin{figure}[ht!]
  \centering
  \includegraphics[width=\linewidth]{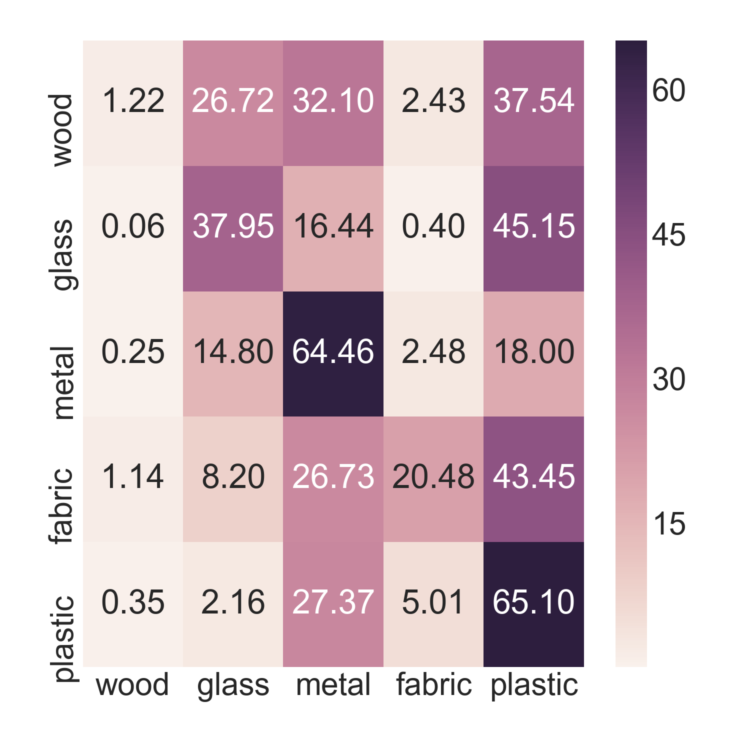}
  \caption{Confusion matrix for top-1 classification predictions for MINC-bw tested on 3D shapes.}
  \label{fig:cm-mincbw}
\end{figure}

To evaluate the baseline of using a 2D material classification network, we use
MINC. The network is based on GoogLeNet. We take their pretrained network and
finetune on their dataset. The classification layer is finetuned to only
classify the five materials we consider in this paper. Furthermore, we choose to
finetune with greyscale images. The reason for this is that our our texture-less
3D renderings do not offer any color cues; therefore, we train the 2D network
under similar conditions. This network is trained until validation losses
converge with batchsize $24$ with stochastic gradient descent with momentum.
The initial learning rate is set to $0.001$ and momentum is set to $0.4$. The
learning rate policy is polynomial decay with power $0.5$. $L_2$ weight decay is
set to $0.0002$. We call this finetuned network MINC-bw. The confusion matrix
for the network on our test 3D renderings is in Fig.~\ref{fig:cm-mincbw}. The
poor performance suggests that it is non-trivial to adapt 2D photos train a
network to learn material descriptors for 3D shapes.



\subsection{\label{sec:tsne}Embedding Visualization}

\begin{figure}[t!]
  \centering
  \includegraphics[width=\linewidth]{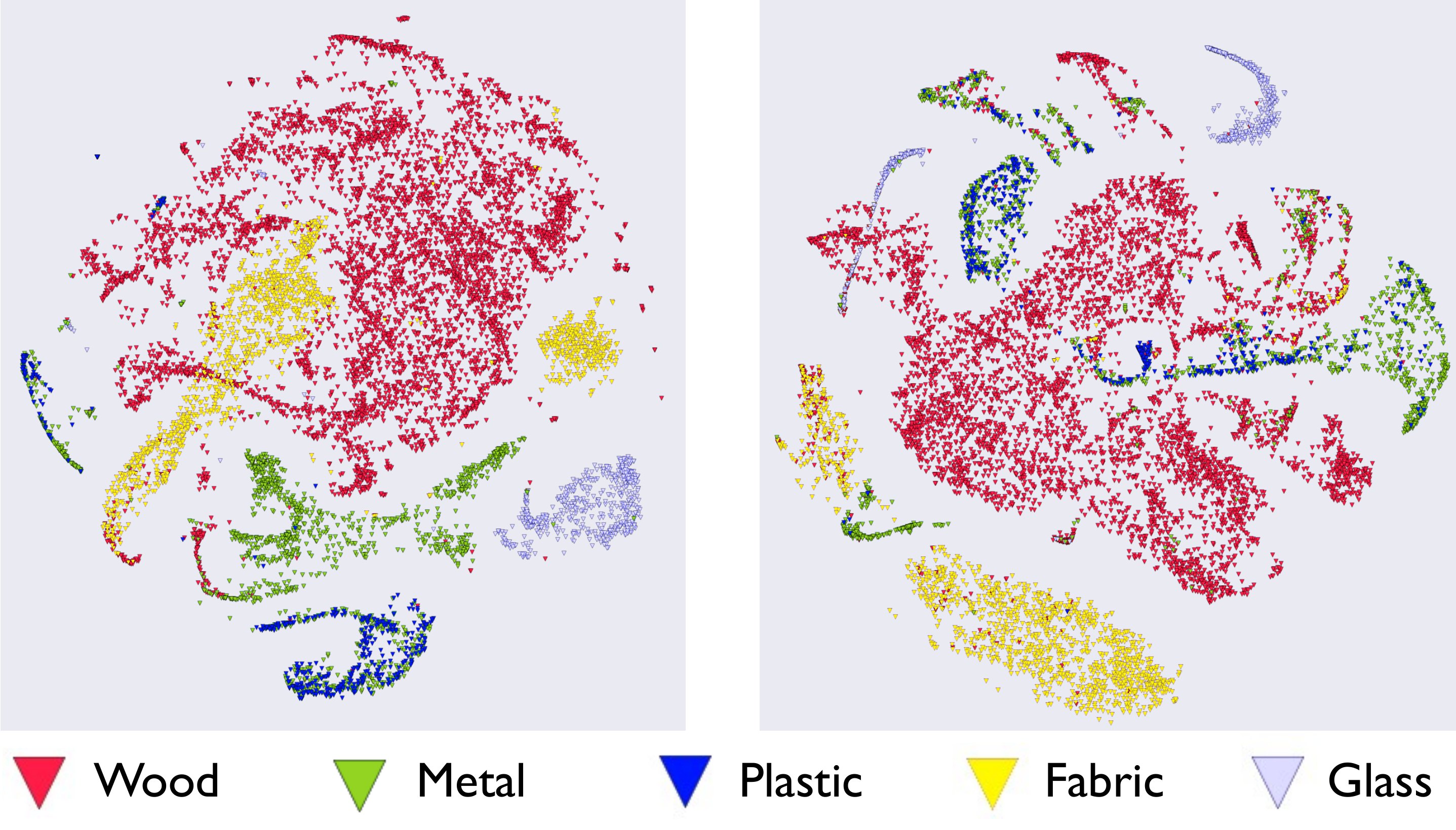}
  \caption{\small t-SNE embeddings for training points. {\em Left:} Classification loss, {\em Right:} Multitask loss. Points with multiple ground truth labels are shown with one label randomly selected.}
  \label{fig:tsne}
\end{figure}

We visualize the learned material-aware descriptor embedding with t-SNE in
\fig{tsne}. In both the Classification and Multitask variations, we see a
tendency of our network to cluster datapoints.

\subsection{\label{sec:cm}Classification vs Multitask Confusion Matrices}

We show the confusion matrices for Classification (as well as Multitask,
for reference) in Fig.~\ref{fig:cm}. Note that Classification predictions are more biased towards
wood. As a result, its glass performance drops after CRF since many glass points
tend to lie on surfaces that resemble wood surfaces (e.g. flat table tops, flat
cabinet doors) -- if many local predictions are wood rather than glass, it is
likely that the CRF will smooth the predictions to wood.

\begin{figure}[ht!]
  \centering
  \begin{subfigure}[b]{0.49\linewidth}
      \includegraphics[width=\linewidth]{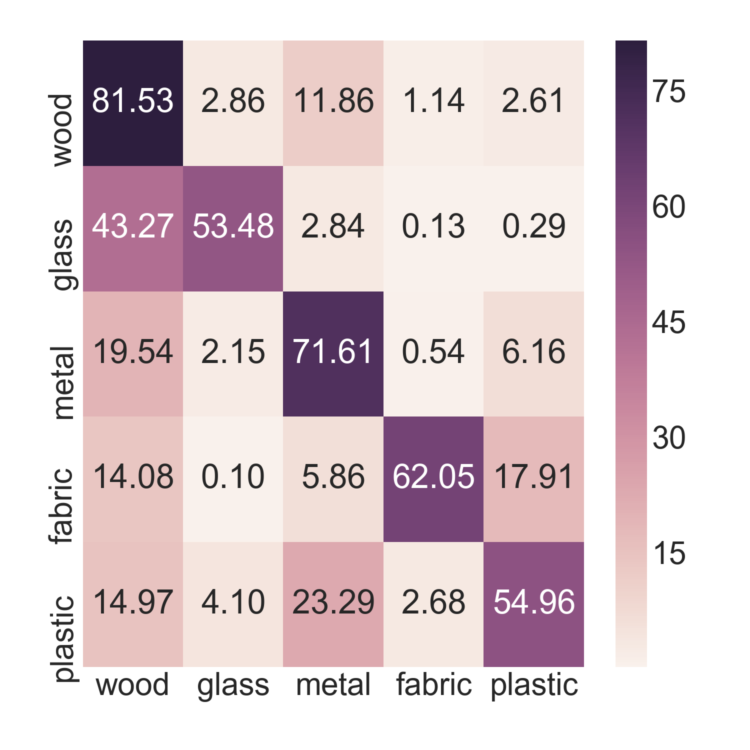}
      \caption{\small Classification}
  \end{subfigure}
  \begin{subfigure}[b]{0.49\linewidth}
      \includegraphics[width=\linewidth]{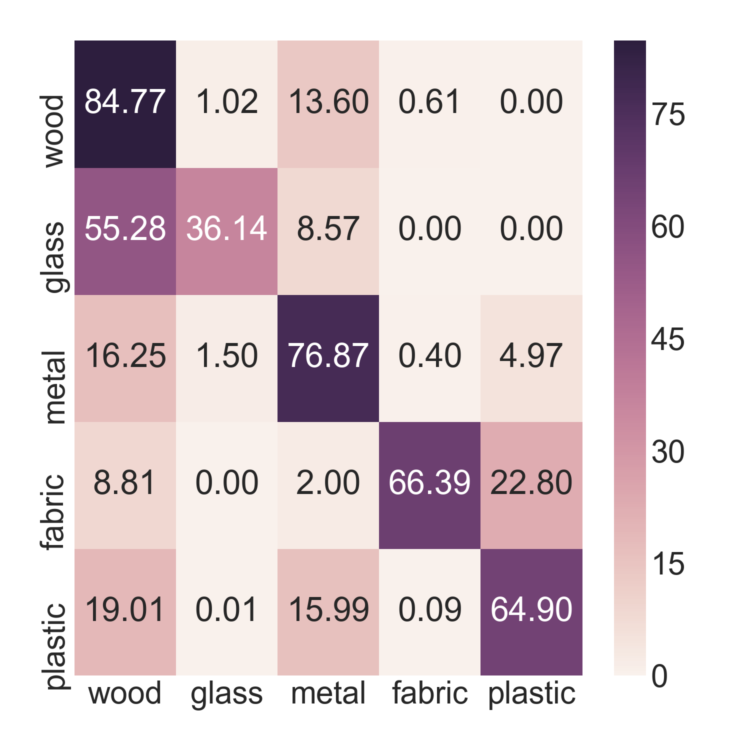}
      \caption{\small Classification + CRF}
  \end{subfigure}
  \begin{subfigure}[b]{0.49\linewidth}
      \includegraphics[width=\linewidth]{figs/cm-15505.png}
      \caption{\small Multitask}
  \end{subfigure}
  \begin{subfigure}[b]{0.49\linewidth}
      \includegraphics[width=\linewidth]{figs/cm-15505-crf.png}
      \caption{\small Multitask + CRF}
  \end{subfigure}
  \caption{\small Confusion matrices for Top-1 classification predictions.}
  \label{fig:cm}
\end{figure}

\subsection{\label{sec:cm-3view}Confusion Matrices for 3 view MVCNN}

\begin{figure}[ht!]
  \centering
  \begin{subfigure}[b]{0.49\linewidth}
      \includegraphics[width=\linewidth]{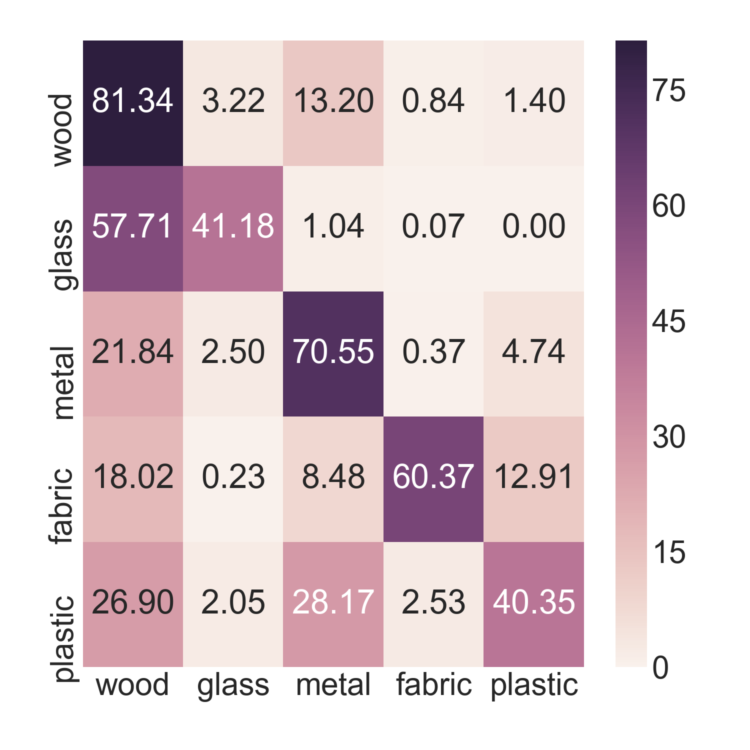}
      \caption{Classification 3 views}
  \end{subfigure}
  \begin{subfigure}[b]{0.49\linewidth}
      \includegraphics[width=\linewidth]{figs/cm-15500.png}
      \caption{Classification 9 views}
  \end{subfigure}
  \begin{subfigure}[b]{0.49\linewidth}
      \includegraphics[width=\linewidth]{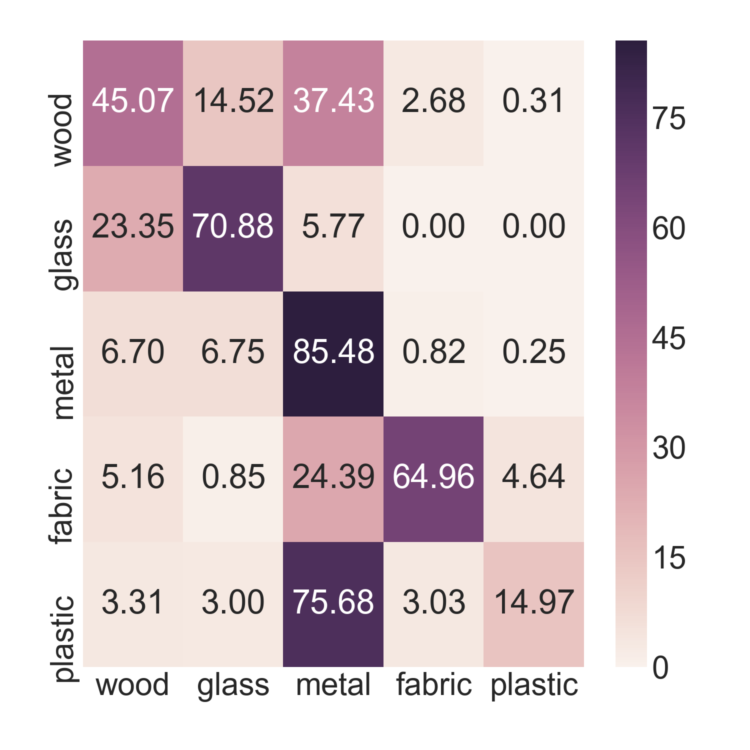}
      \caption{Multitask 3 views}
  \end{subfigure}
  \begin{subfigure}[b]{0.49\linewidth}
      \includegraphics[width=\linewidth]{figs/cm-15505.png}
      \caption{Multitask 9 views}
  \end{subfigure}
  \caption{Confusion matrices for Top-1 classification predictions.}
  \label{fig:cm_reduced_view}
\end{figure}

The confusion matrices for 3 view MVCNNs (1 viewpoint, 3 distances) are in
Fig.~\ref{fig:cm_reduced_view}. For reference, the matrices for the 9 view
MVCNNs (3 viewpoints, 3 distances) are also shown. Note that confusions are
reduced when using 9 views over 3 views. For both Classification and Multitask,
fabric performance is relatively unaffected by reduced views while plastic
suffers. In Classification 3 views, wood predictions dominate. In Multitask 3
views, metal predictions dominate -- as a consequence, glass does relatively well
(since glass is typically competing with wood) and plastic does extremely poorly
(since plastic parts can often be shaped like metal and our training dataset
contains a high number of ``plastic or metal`` labels relative to ``plastic''
labels).

\subsection{\label{sec:cm-contrastive}Confusion Matrix for Contrastive Loss Only}

\begin{figure}[ht!]
  \centering
  \includegraphics[width=\linewidth]{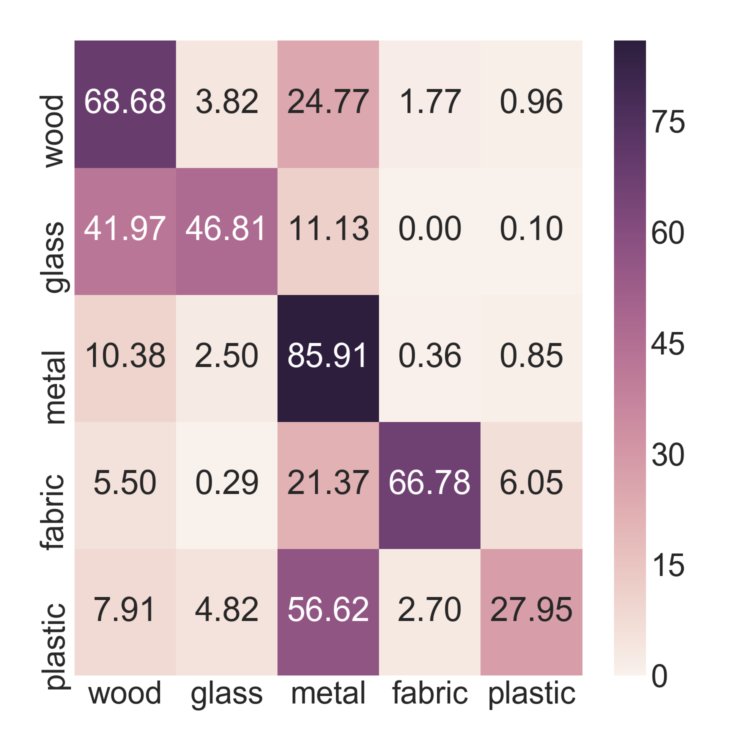}
  \caption{Confusion matrix for Top-1 classification predictions for network
    trained with contrastive loss only.}
  \label{fig:cm-contrastive}
\end{figure}

The MVCNN trained with only contrastive loss achieves a mean class top 1
accuracy of 59\% (in comparison to 65\% with Classification and 66\% with
Multitask).  This variant often confuses plastic for metal, and performs
poorly on glass relative to the Classification or Multitask variants. The confusion matrix is shown in Fig.~\ref{fig:cm-contrastive}.

\subsection{\label{sec:data-vis}Sample of Dataset and Predictions}

Here we show some samples from both our high-quality expert-annotated benchmark
dataset as well as our large crowdsourced training dataset. Please refer to the
legend by each shape for labels. The colors are consistent within each figure
but may not be across figures.

\begin{figure*}[ht!]
  \centering
  \includegraphics[width=0.23\linewidth]{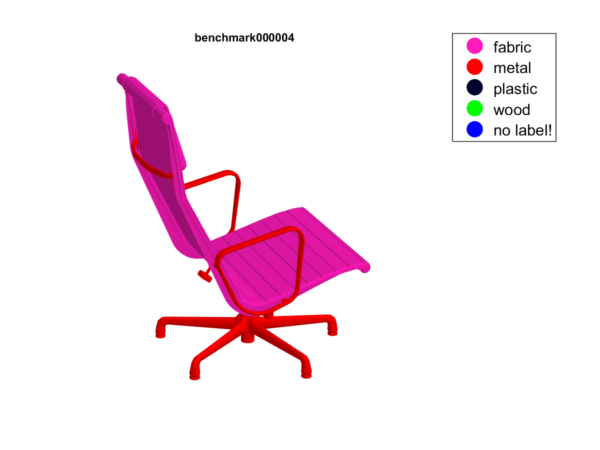}
  \includegraphics[width=0.23\linewidth]{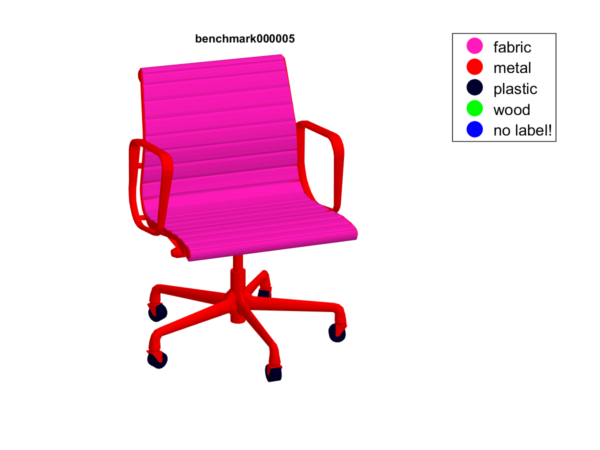}
  \includegraphics[width=0.23\linewidth]{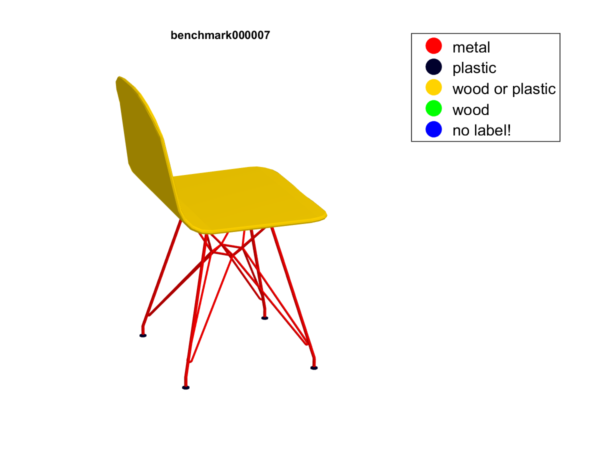}
  \includegraphics[width=0.23\linewidth]{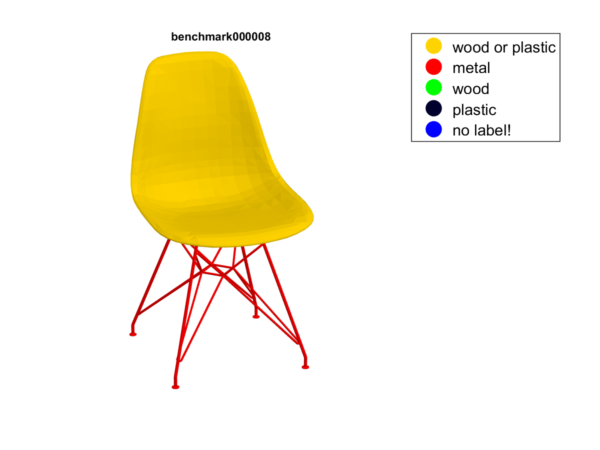}
  \includegraphics[width=0.23\linewidth]{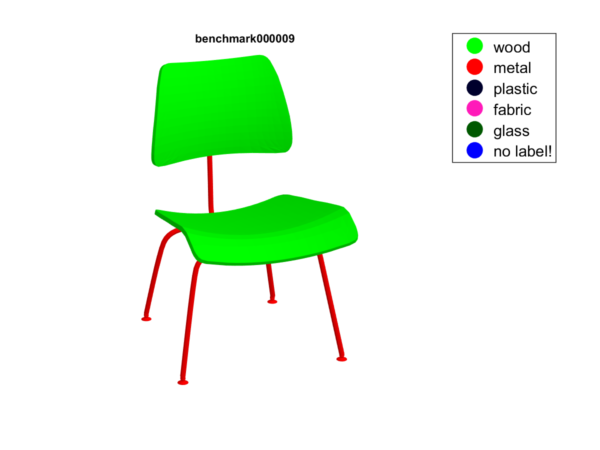}
  \includegraphics[width=0.23\linewidth]{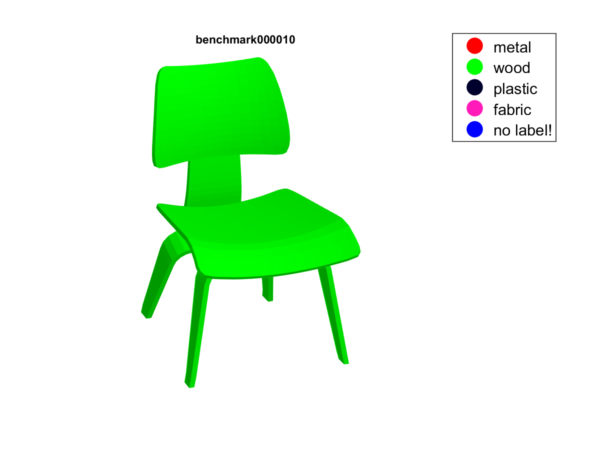}
  \includegraphics[width=0.23\linewidth]{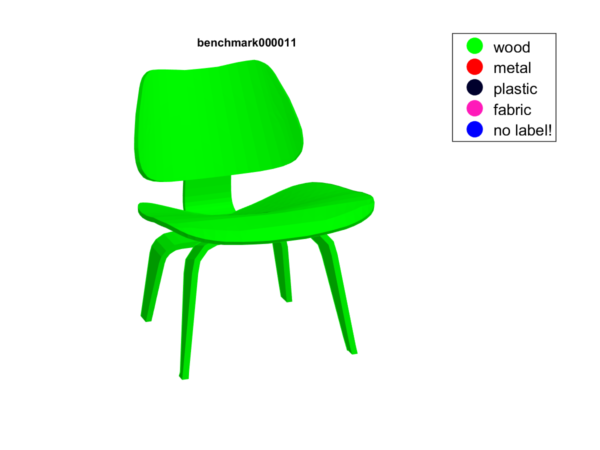}
  \includegraphics[width=0.23\linewidth]{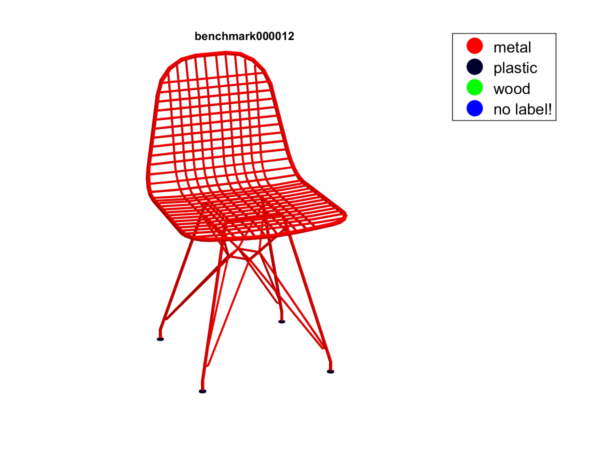}
  \includegraphics[width=0.23\linewidth]{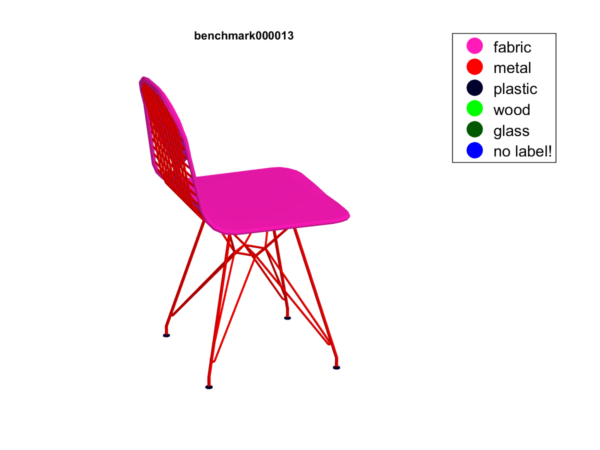}
  \includegraphics[width=0.23\linewidth]{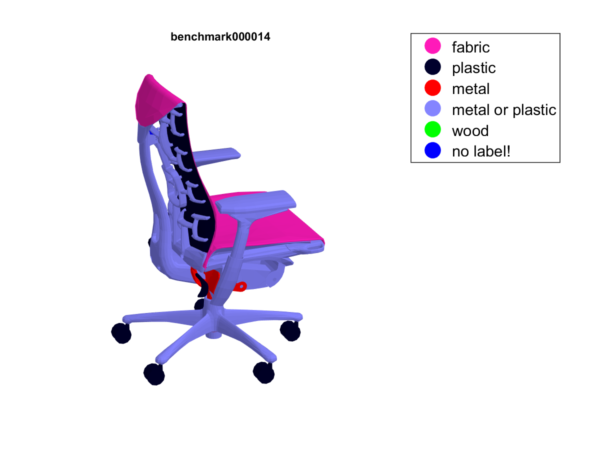}
  \includegraphics[width=0.23\linewidth]{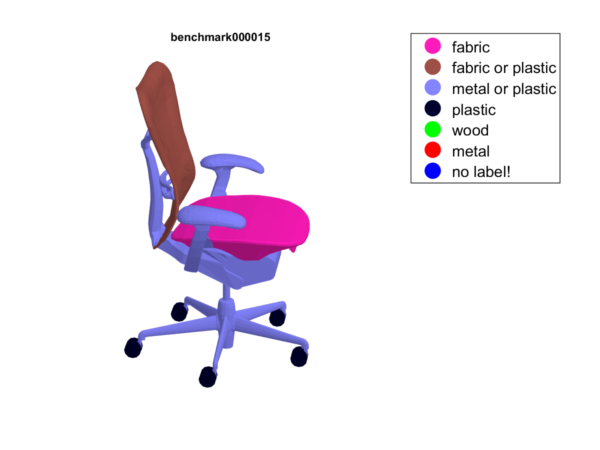}
  \includegraphics[width=0.23\linewidth]{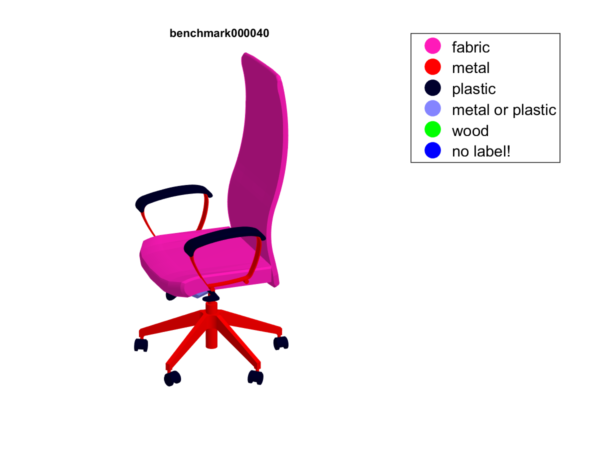}
  \includegraphics[width=0.23\linewidth]{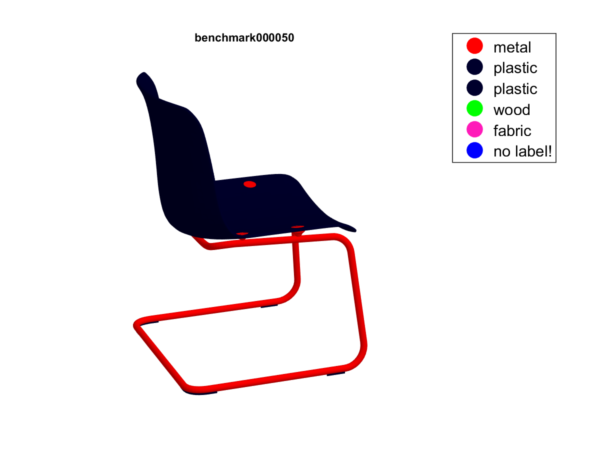}
  \includegraphics[width=0.23\linewidth]{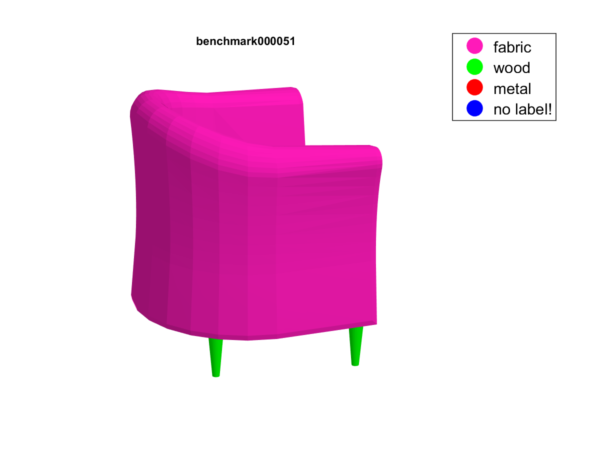}
  \includegraphics[width=0.23\linewidth]{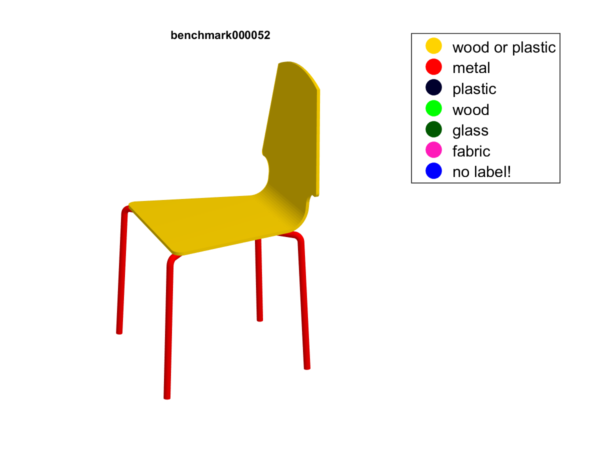}
  \includegraphics[width=0.23\linewidth]{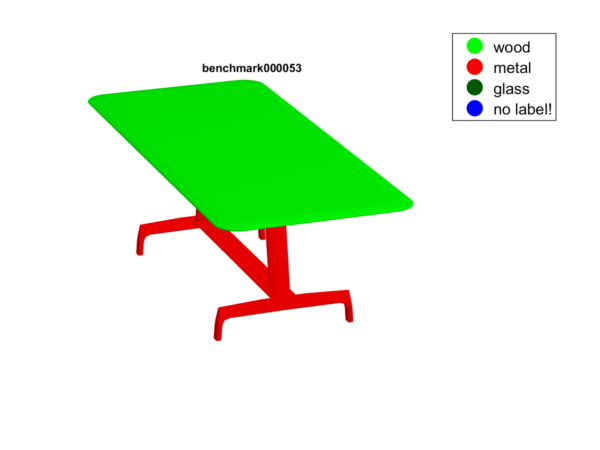}
  \includegraphics[width=0.23\linewidth]{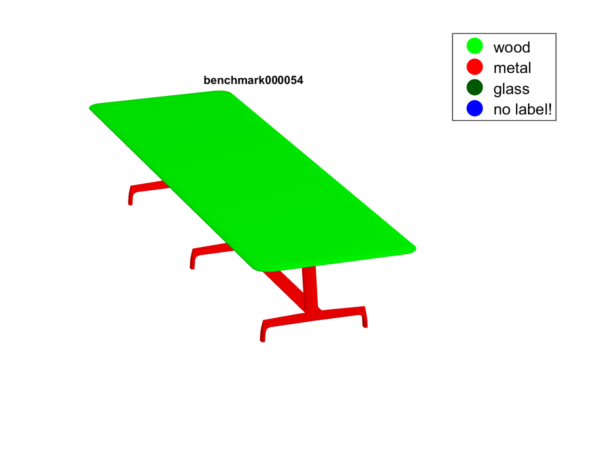}
  \includegraphics[width=0.23\linewidth]{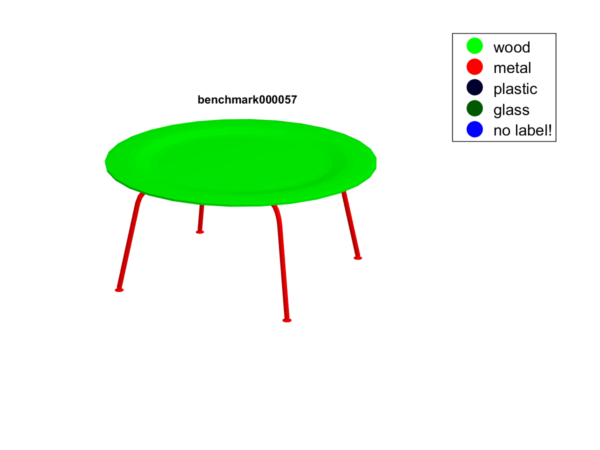}
  \includegraphics[width=0.23\linewidth]{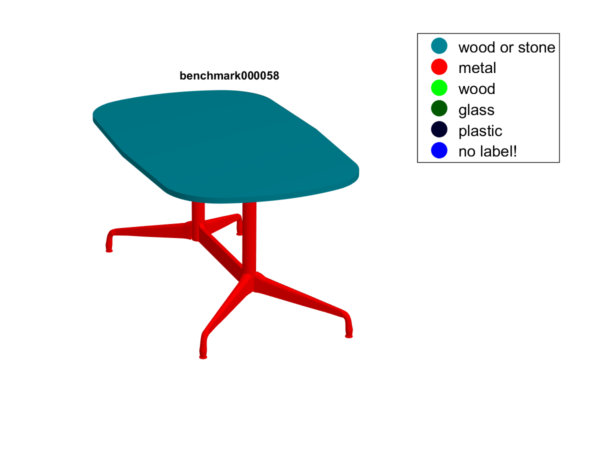}
  \includegraphics[width=0.23\linewidth]{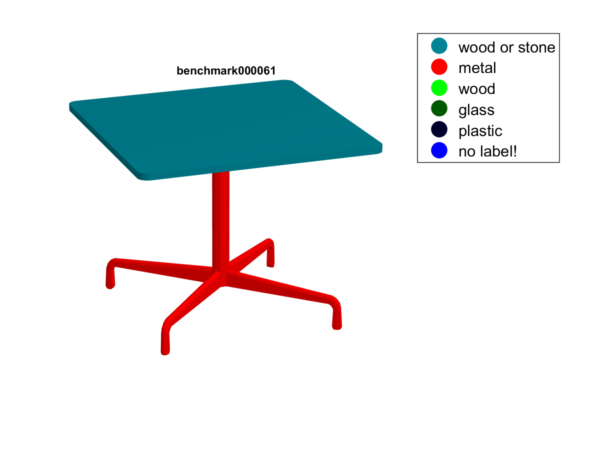}
  \includegraphics[width=0.23\linewidth]{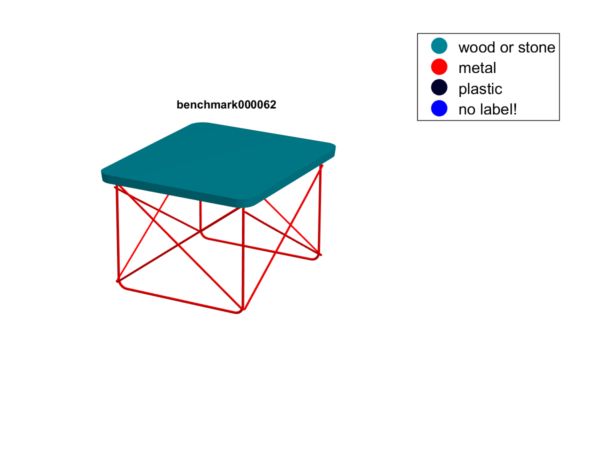}
  \includegraphics[width=0.23\linewidth]{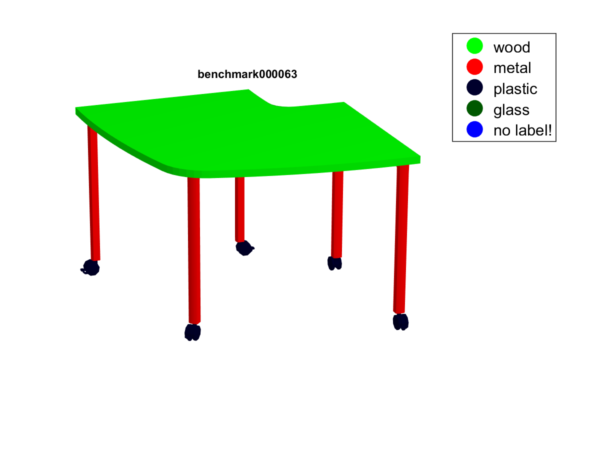}
  \includegraphics[width=0.23\linewidth]{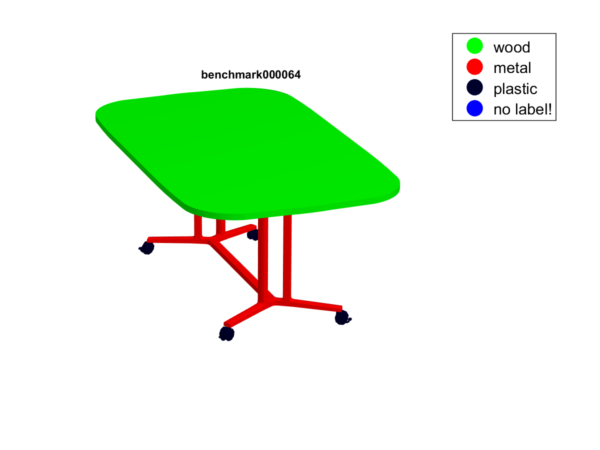}
  \includegraphics[width=0.23\linewidth]{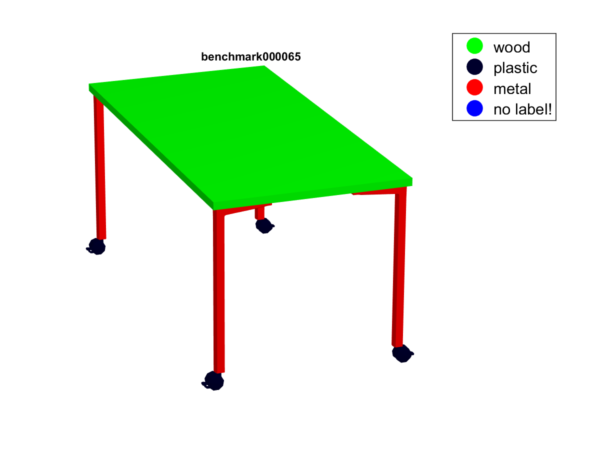}
  \includegraphics[width=0.23\linewidth]{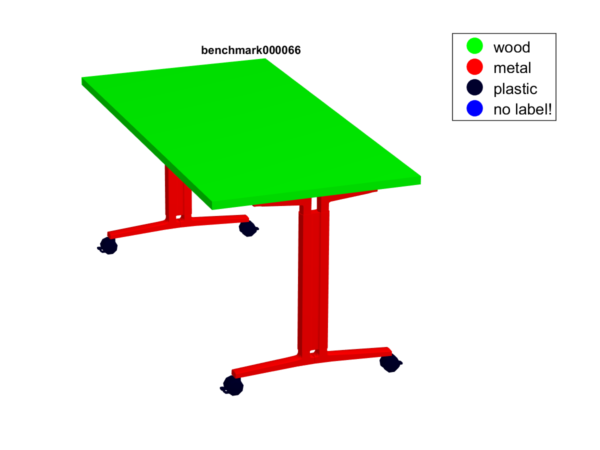}
  \includegraphics[width=0.23\linewidth]{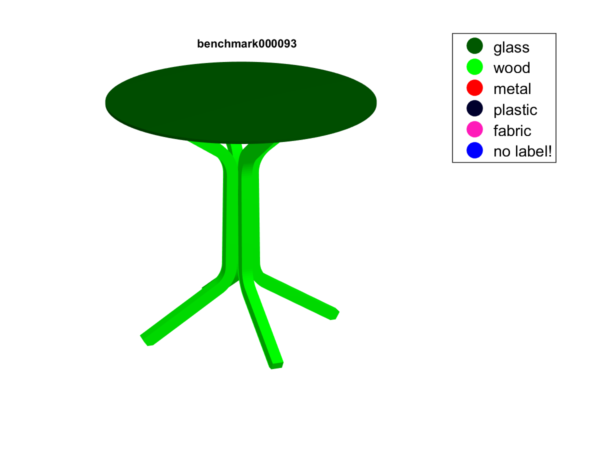}
  \includegraphics[width=0.23\linewidth]{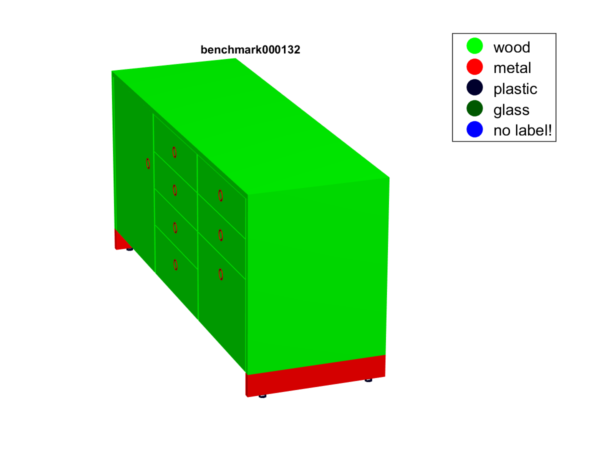}
  \includegraphics[width=0.23\linewidth]{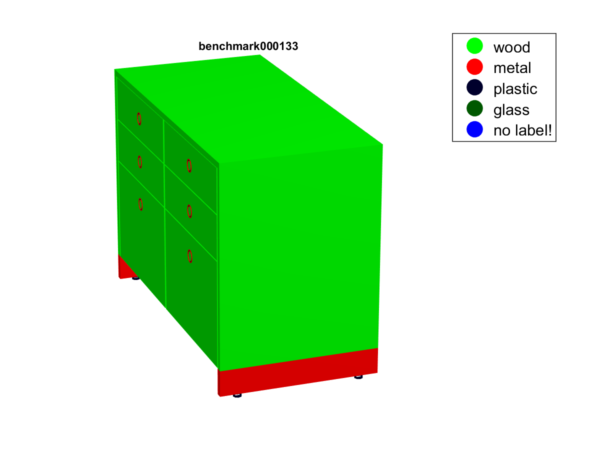}
  \vspace{-1mm}
  \caption{\small Small selection of benchmark shapes. Ground truth labels
  are shown. Please refer to legend for each shape for labels.}
  \label{fig:benchmarkgt}
  \vspace{-2mm}
\end{figure*}

\begin{figure*}[ht!]
  \centering
  \includegraphics[width=0.23\linewidth]{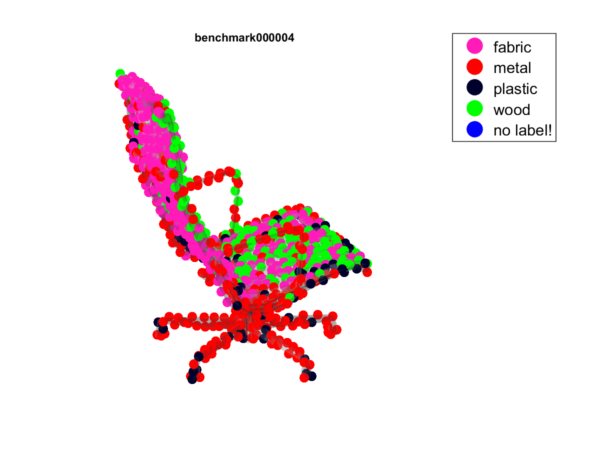}
  \includegraphics[width=0.23\linewidth]{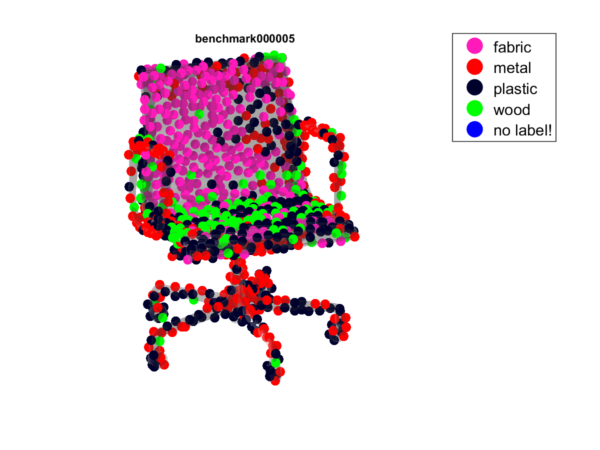}
  \includegraphics[width=0.23\linewidth]{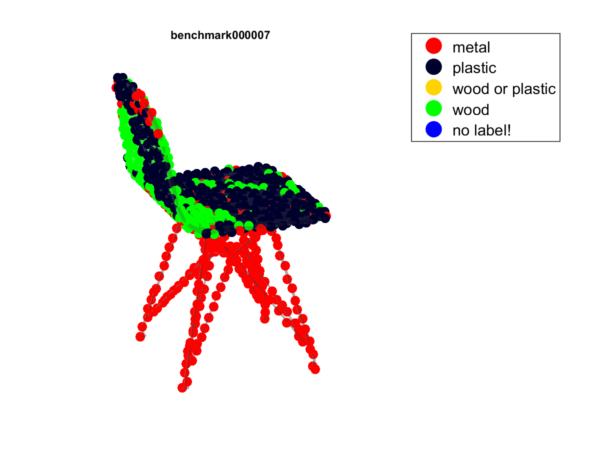}
  \includegraphics[width=0.23\linewidth]{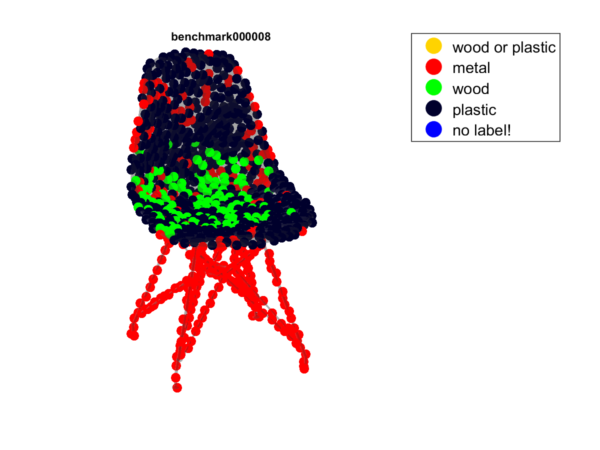}
  \includegraphics[width=0.23\linewidth]{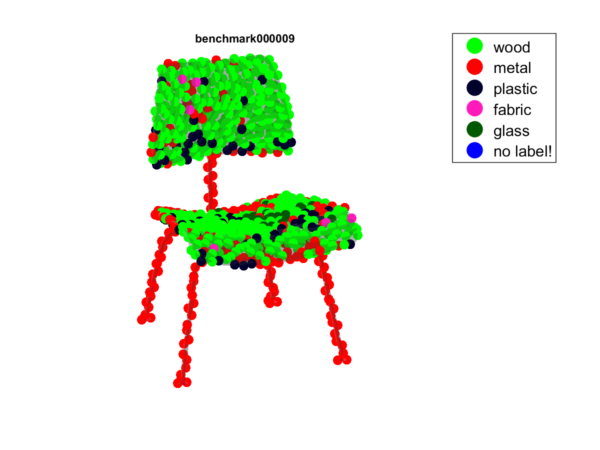}
  \includegraphics[width=0.23\linewidth]{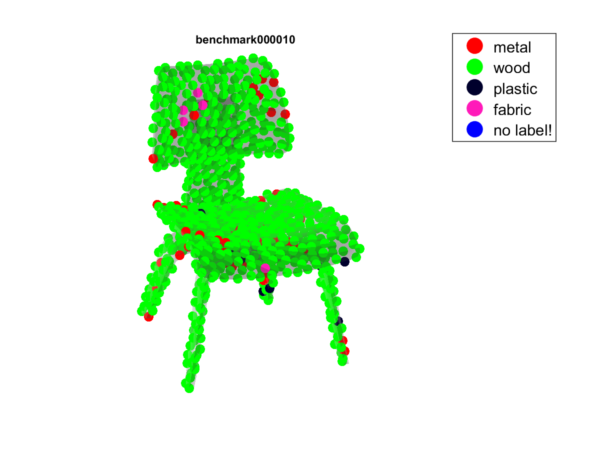}
  \includegraphics[width=0.23\linewidth]{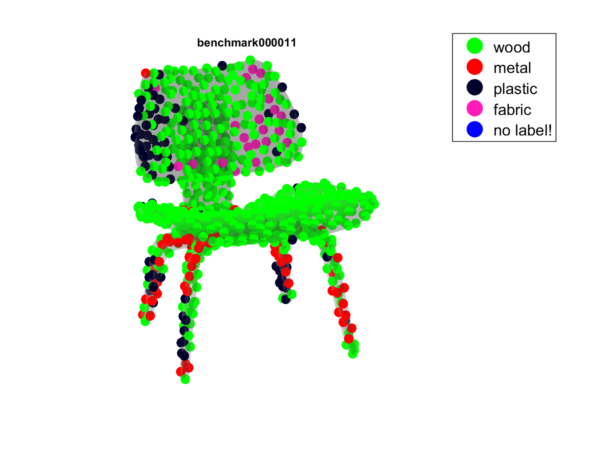}
  \includegraphics[width=0.23\linewidth]{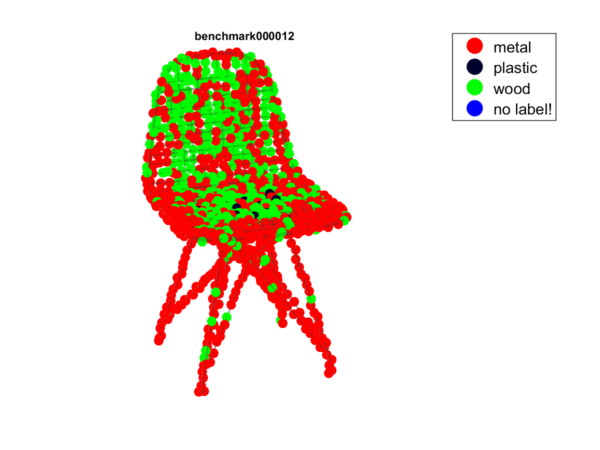}
  \includegraphics[width=0.23\linewidth]{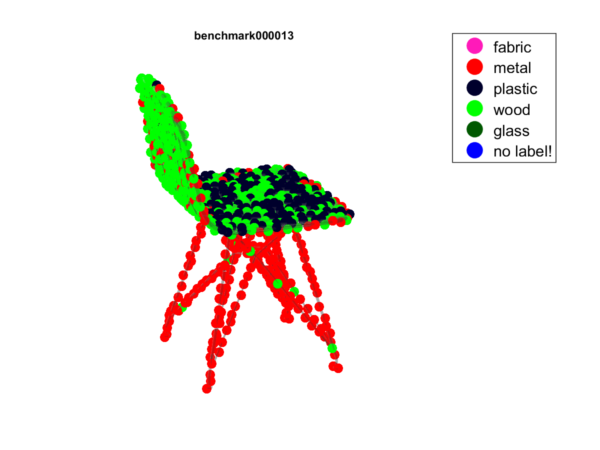}
  \includegraphics[width=0.23\linewidth]{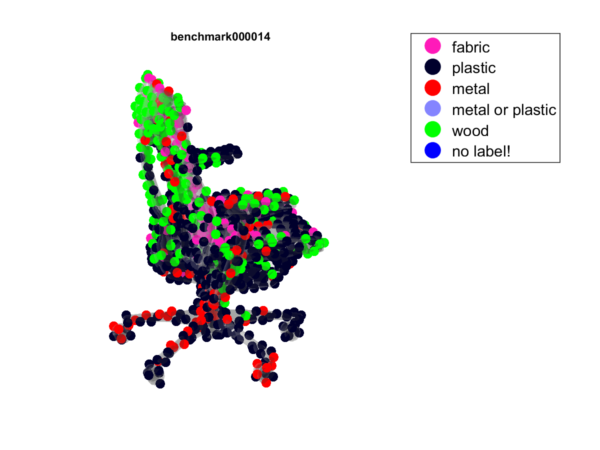}
  \includegraphics[width=0.23\linewidth]{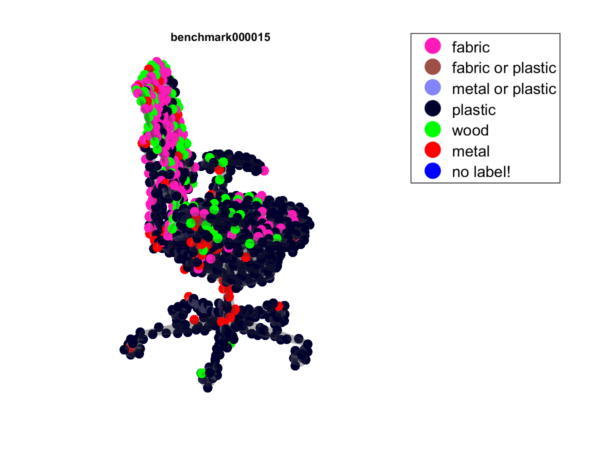}
  \includegraphics[width=0.23\linewidth]{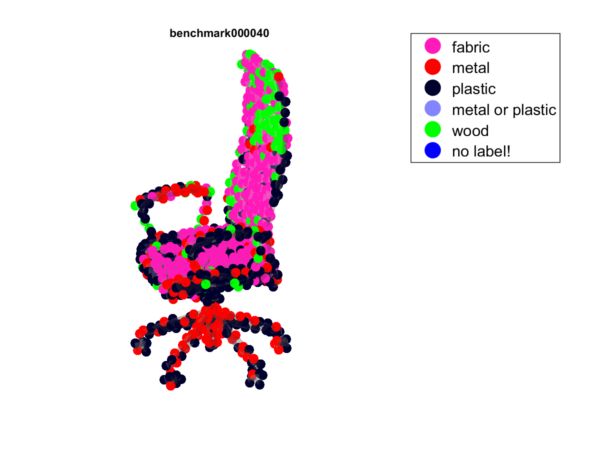}
  \includegraphics[width=0.23\linewidth]{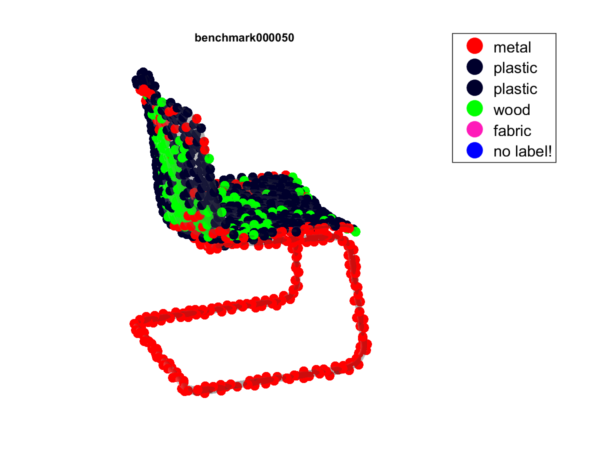}
  \includegraphics[width=0.23\linewidth]{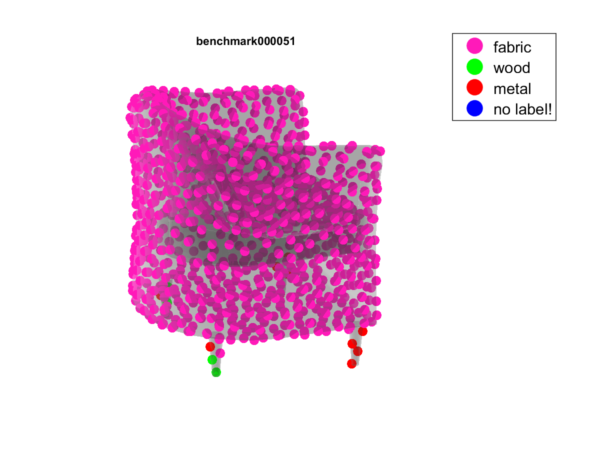}
  \includegraphics[width=0.23\linewidth]{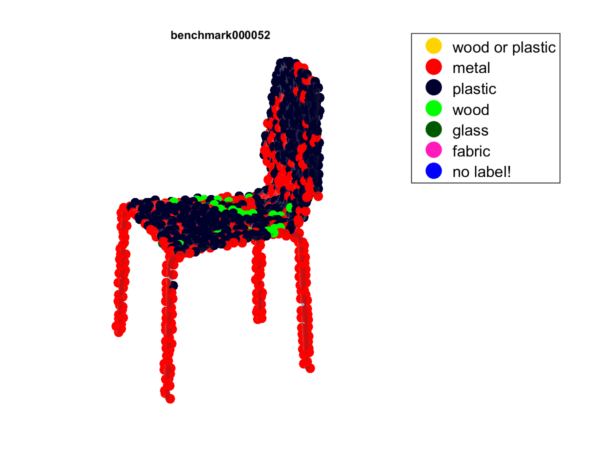}
  \includegraphics[width=0.23\linewidth]{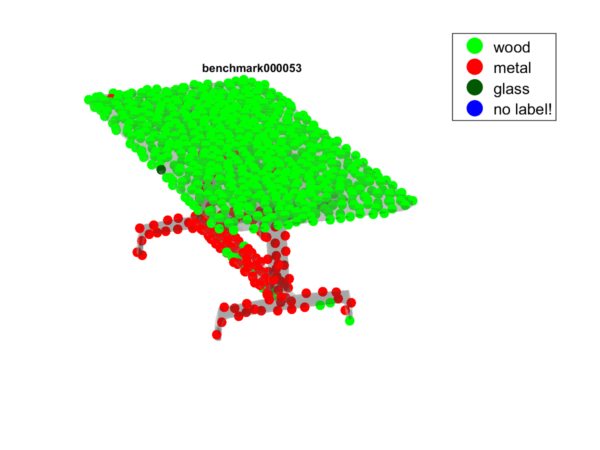}
  \includegraphics[width=0.23\linewidth]{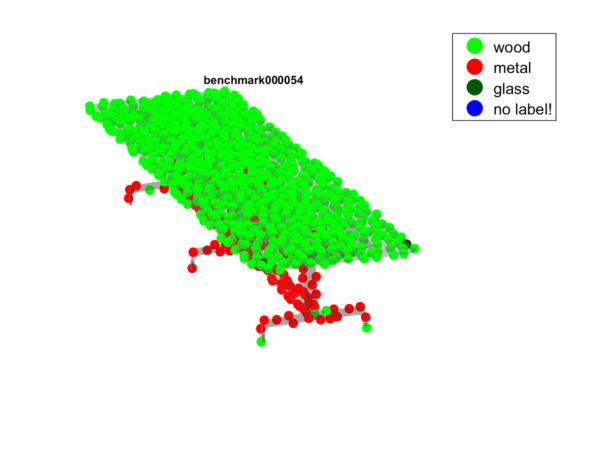}
  \includegraphics[width=0.23\linewidth]{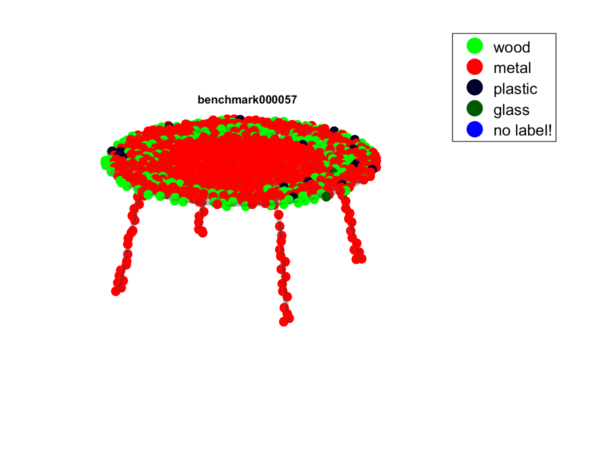}
  \includegraphics[width=0.23\linewidth]{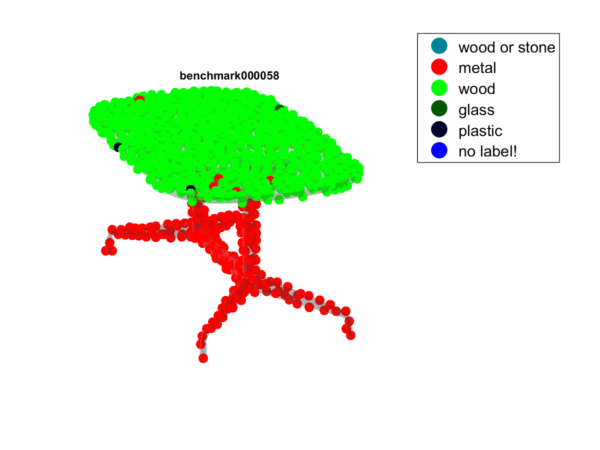}
  \includegraphics[width=0.23\linewidth]{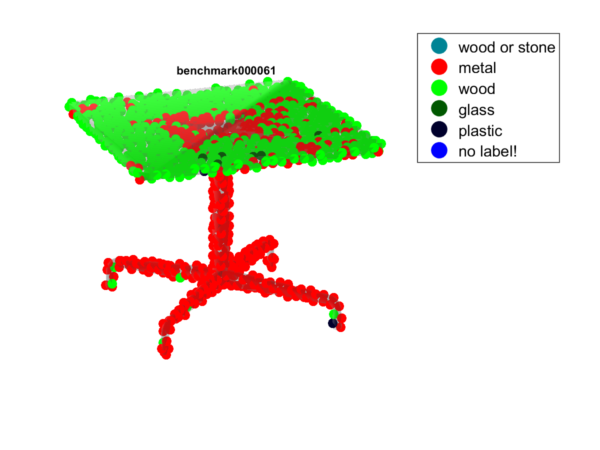}
  \includegraphics[width=0.23\linewidth]{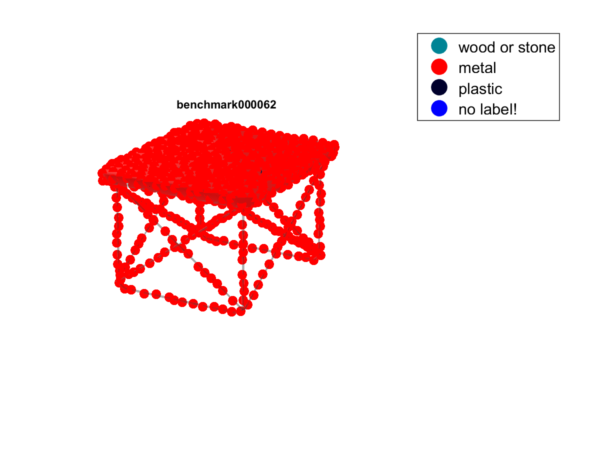}
  \includegraphics[width=0.23\linewidth]{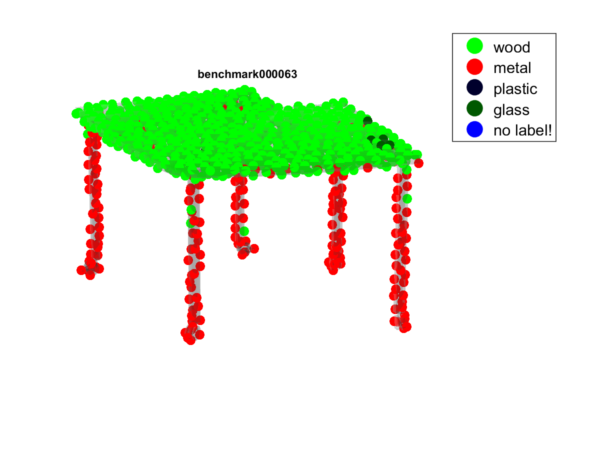}
  \includegraphics[width=0.23\linewidth]{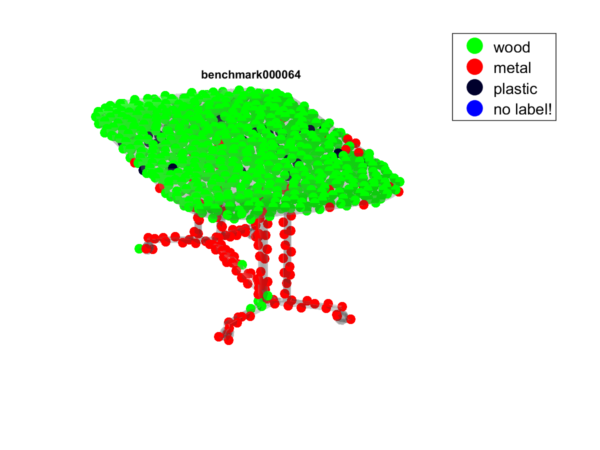}
  \includegraphics[width=0.23\linewidth]{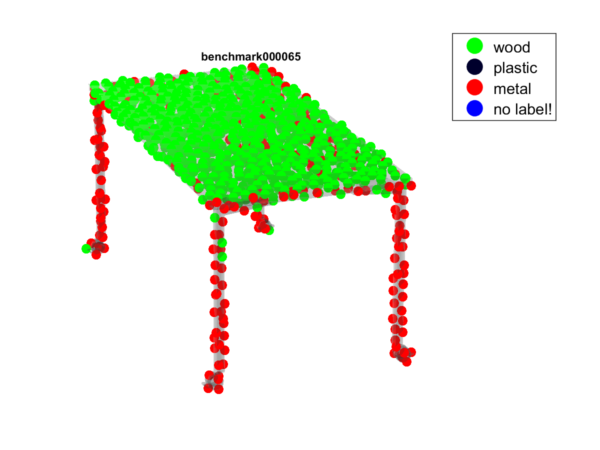}
  \includegraphics[width=0.23\linewidth]{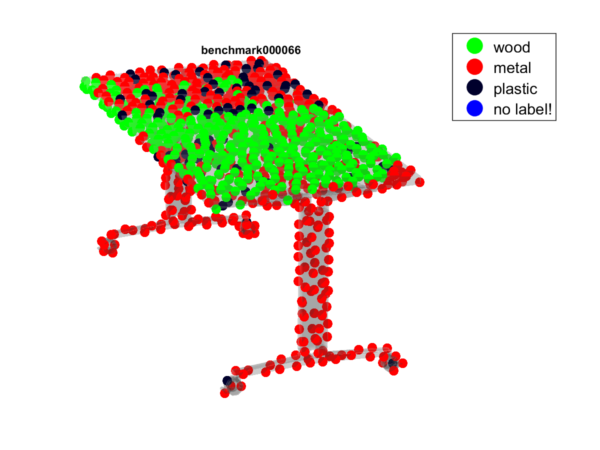}
  \includegraphics[width=0.23\linewidth]{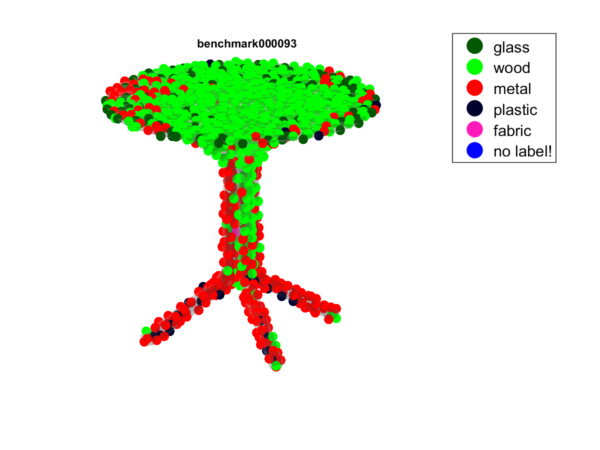}
  \includegraphics[width=0.23\linewidth]{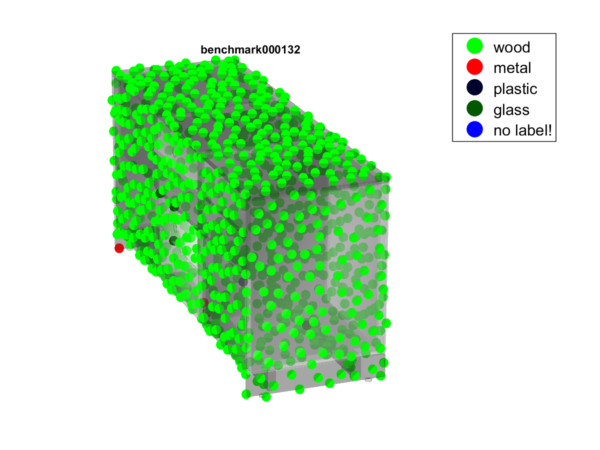}
  \includegraphics[width=0.23\linewidth]{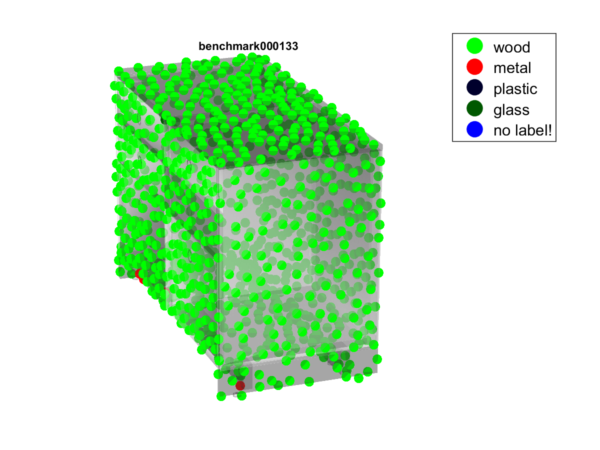}
  \vspace{-1mm}
  \caption{\small MVCNN point-predictions on benchmark shapes. Please refer to
  legend for each shape for labels. }
  \label{fig:benchmarkpointpred}
  \vspace{-2mm}
\end{figure*}

\begin{figure*}[ht!]
  \centering
  \includegraphics[width=0.23\linewidth]{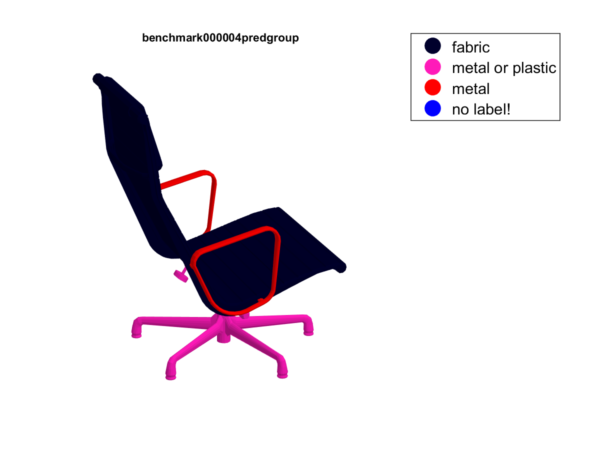}
  \includegraphics[width=0.23\linewidth]{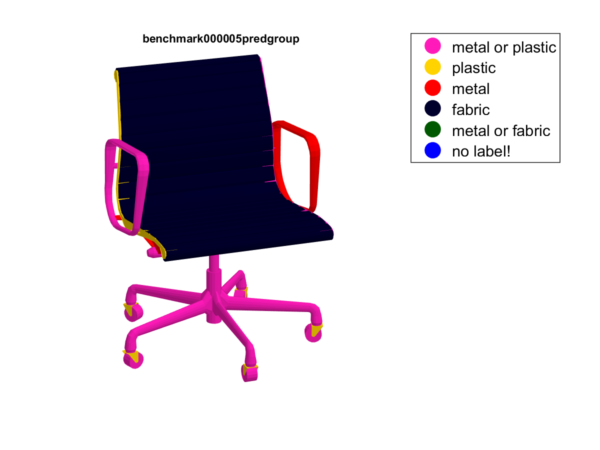}
  \includegraphics[width=0.23\linewidth]{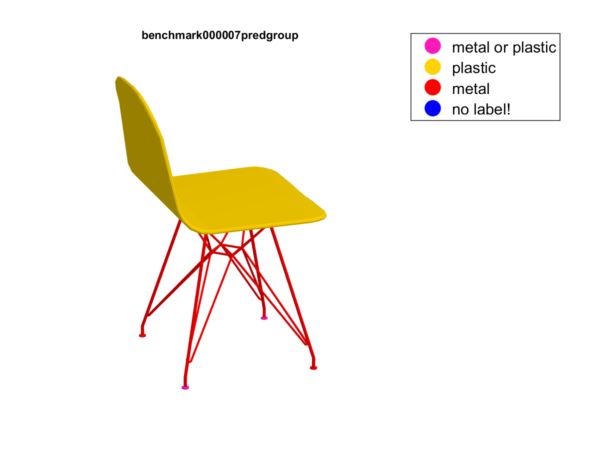}
  \includegraphics[width=0.23\linewidth]{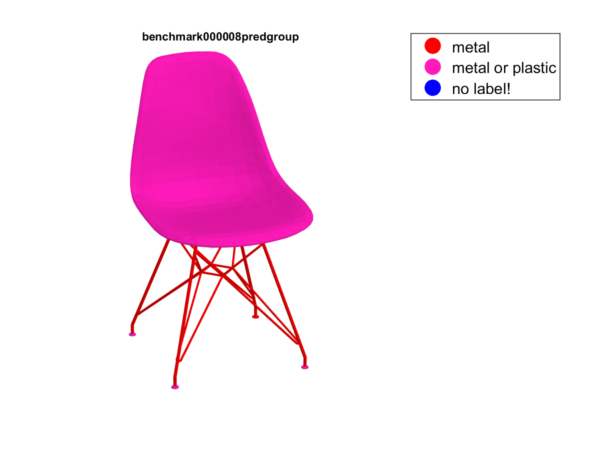}
  \includegraphics[width=0.23\linewidth]{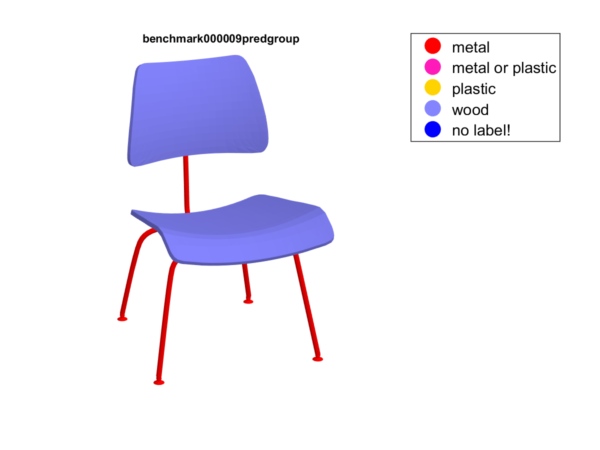}
  \includegraphics[width=0.23\linewidth]{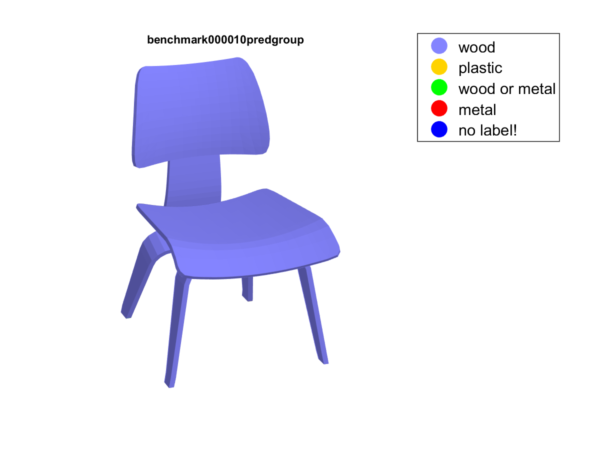}
  \includegraphics[width=0.23\linewidth]{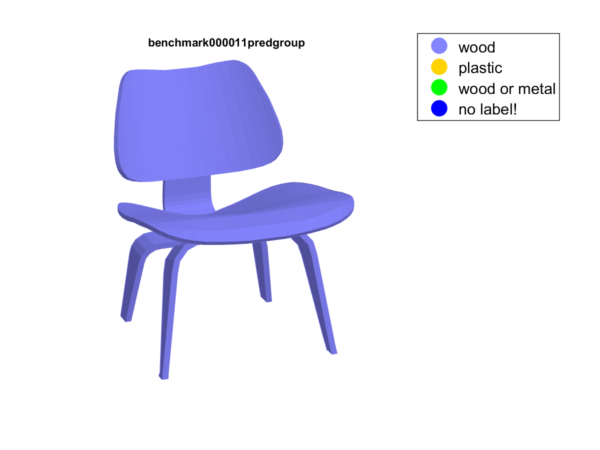}
  \includegraphics[width=0.23\linewidth]{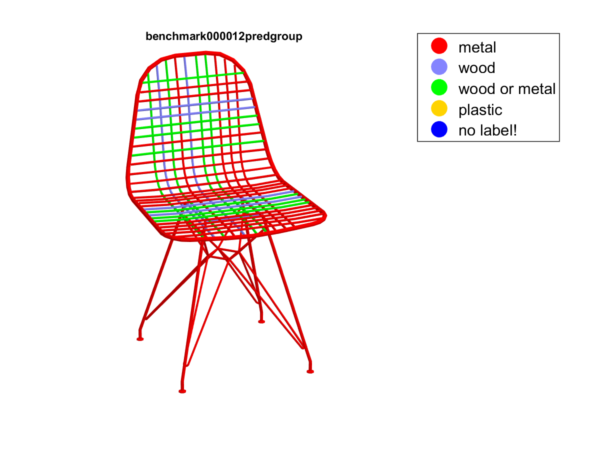}
  \includegraphics[width=0.23\linewidth]{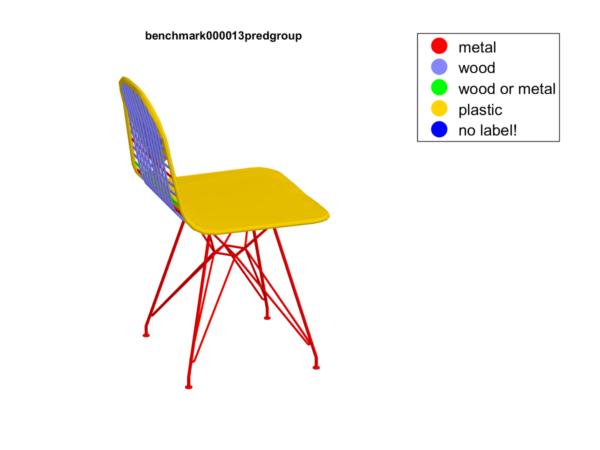}
  \includegraphics[width=0.23\linewidth]{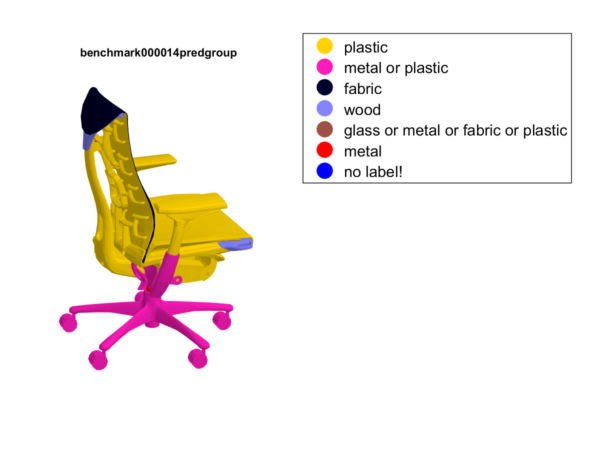}
  \includegraphics[width=0.23\linewidth]{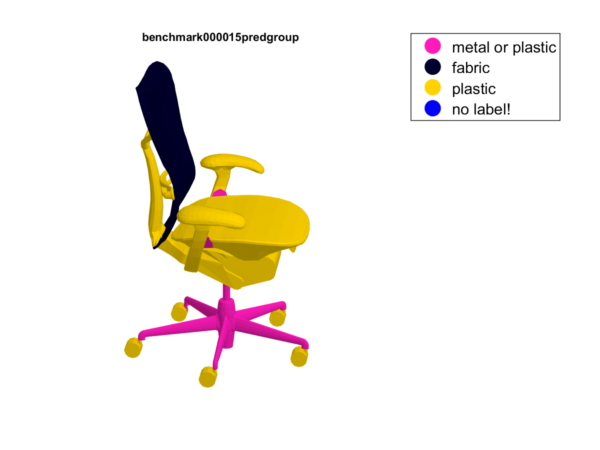}
  \includegraphics[width=0.23\linewidth]{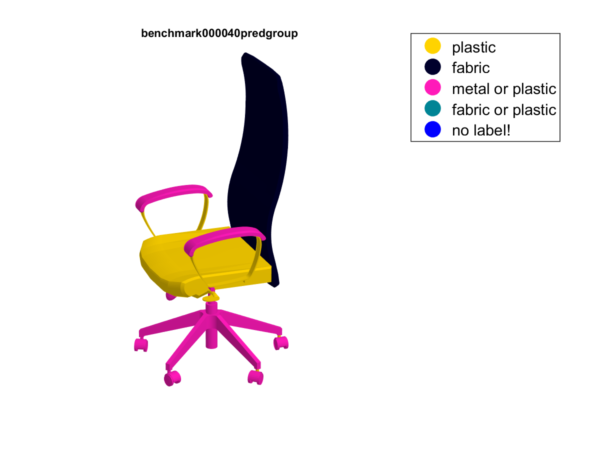}
  \includegraphics[width=0.23\linewidth]{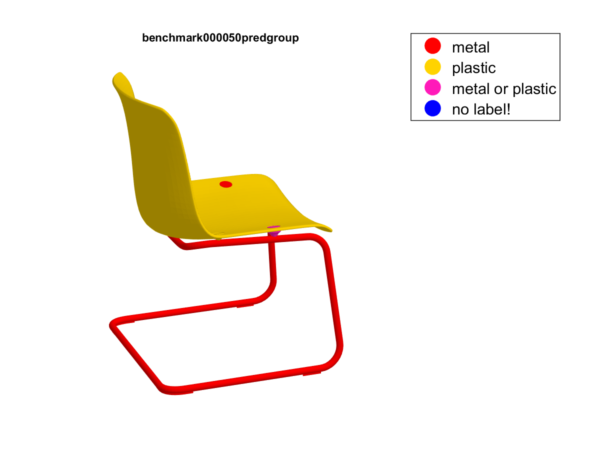}
  \includegraphics[width=0.23\linewidth]{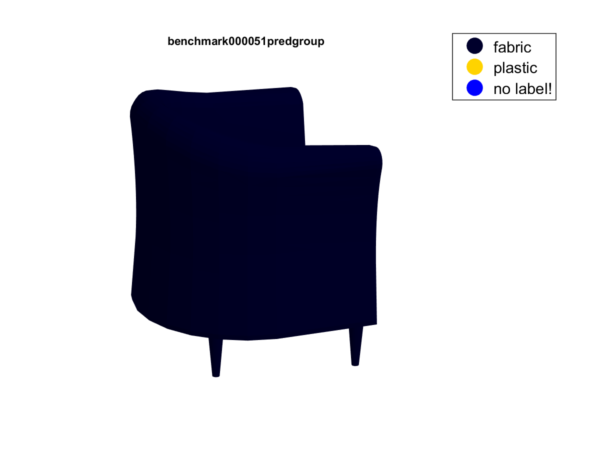}
  \includegraphics[width=0.23\linewidth]{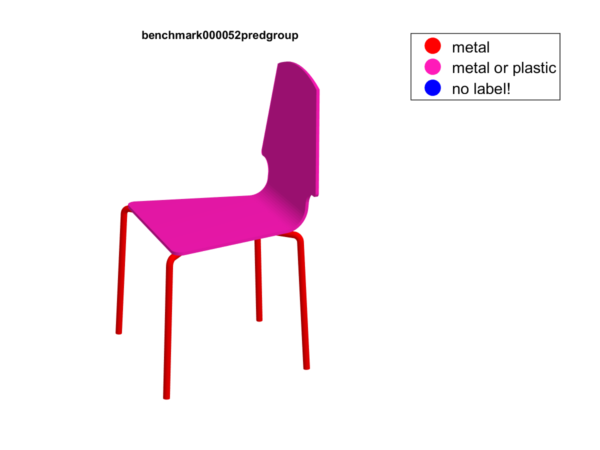}
  \includegraphics[width=0.23\linewidth]{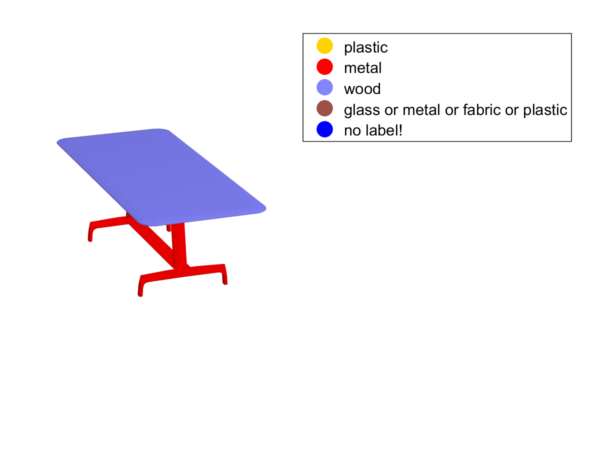}
  \includegraphics[width=0.23\linewidth]{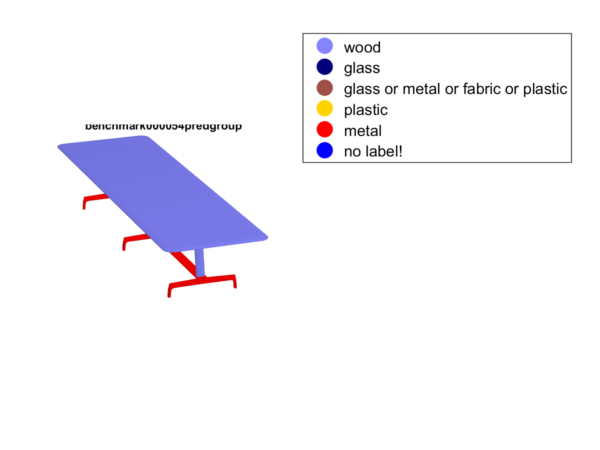}
  \includegraphics[width=0.23\linewidth]{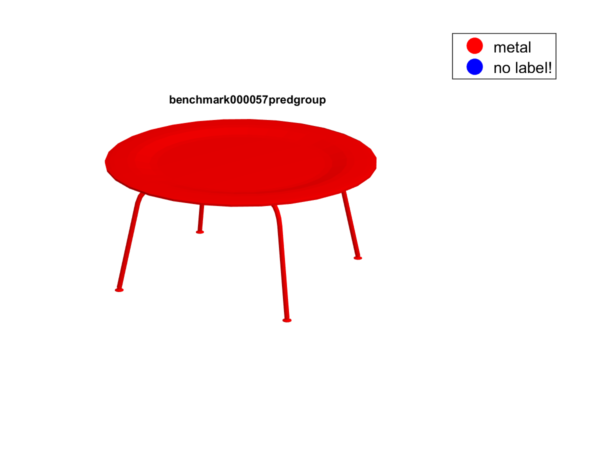}
  \includegraphics[width=0.23\linewidth]{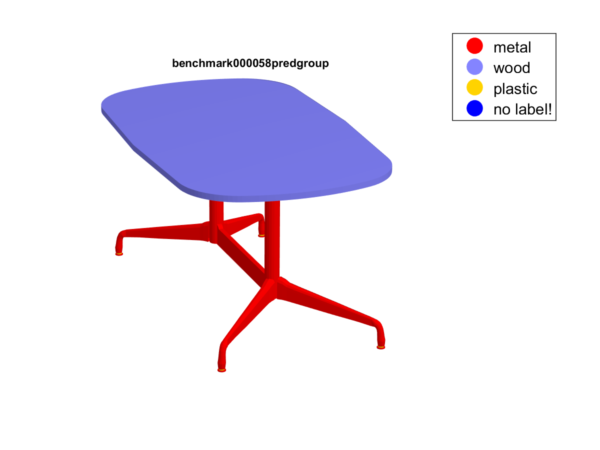}
  \includegraphics[width=0.23\linewidth]{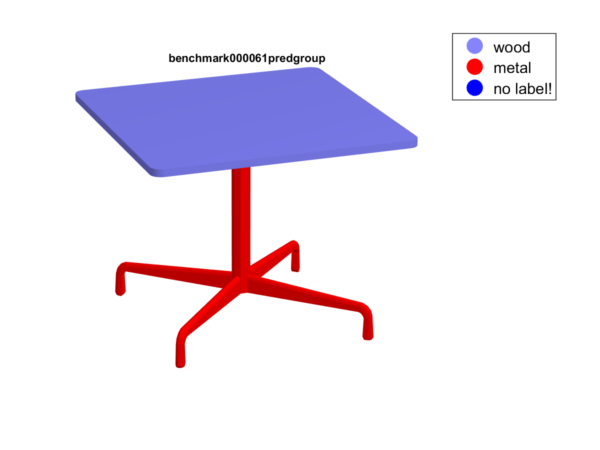}
  \includegraphics[width=0.23\linewidth]{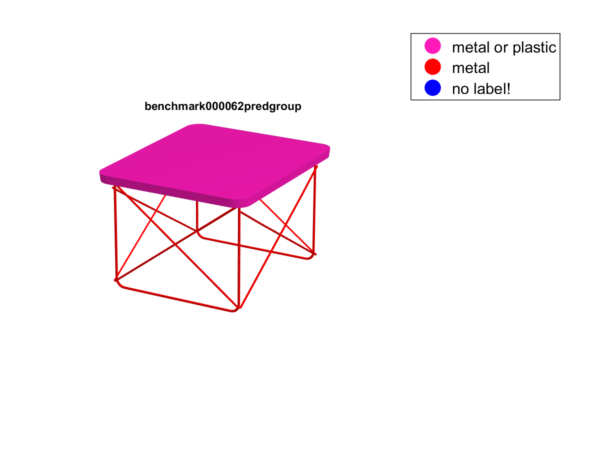}
  \includegraphics[width=0.23\linewidth]{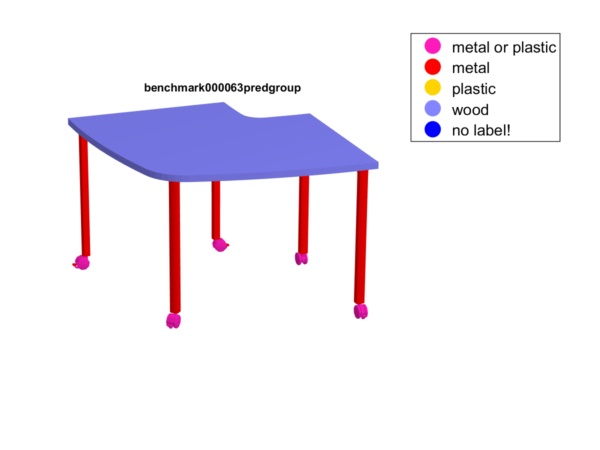}
  \includegraphics[width=0.23\linewidth]{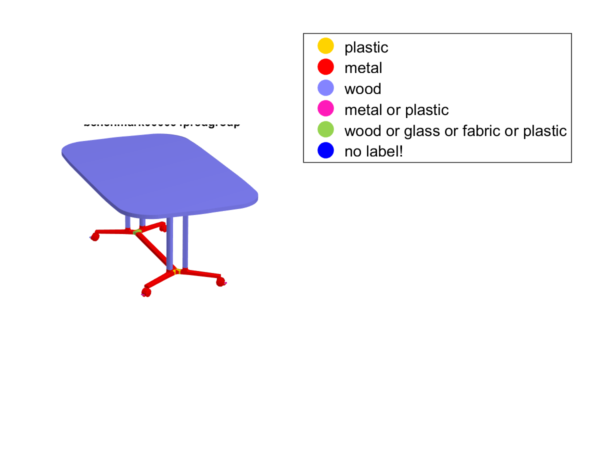}
  \includegraphics[width=0.23\linewidth]{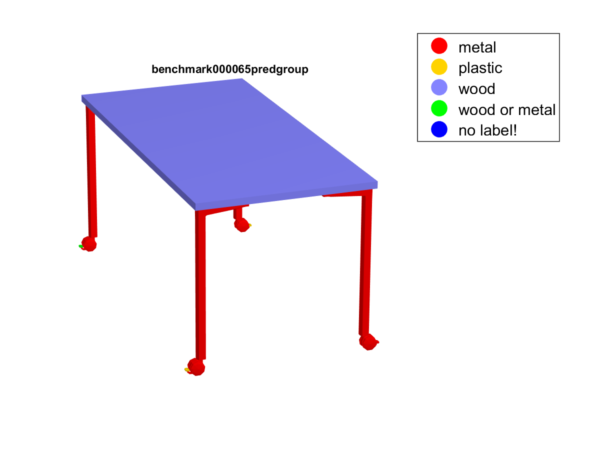}
  \includegraphics[width=0.23\linewidth]{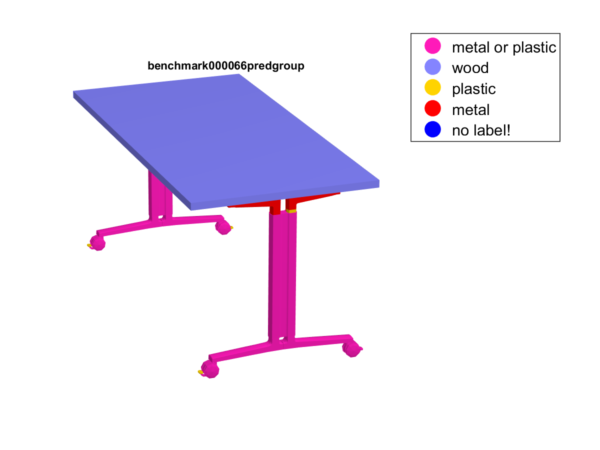}
  \includegraphics[width=0.23\linewidth]{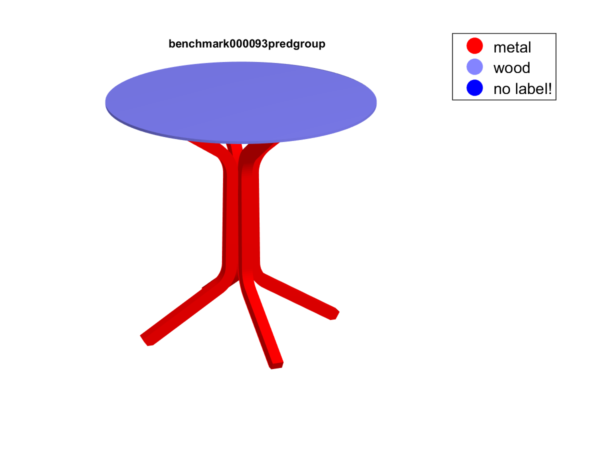}
  \includegraphics[width=0.23\linewidth]{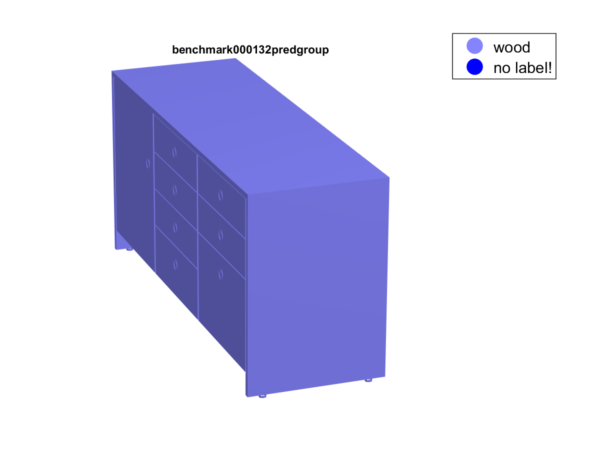}
  \includegraphics[width=0.23\linewidth]{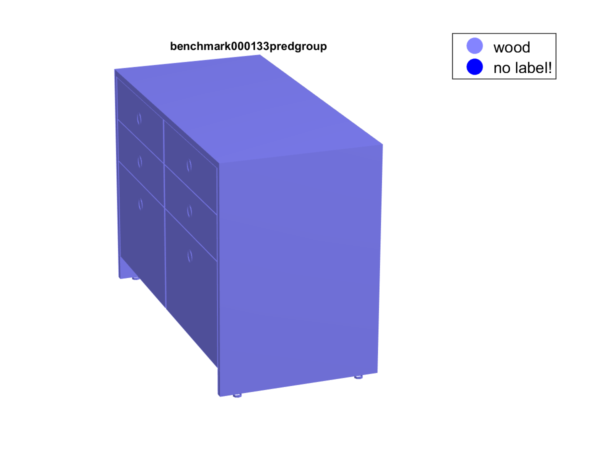}
  \vspace{-1mm}
  \caption{\small MVCNN+CRF part-predictions on benchmark shapes. Please refer to legend for
each shape for labels. (Note that colors are not consistent with
Figs ~\ref{fig:benchmarkgt},~\ref{fig:benchmarkpointpred}) }
  \label{fig:benchmarkpartpred}
  \vspace{-2mm}
\end{figure*}

\begin{figure*}[ht!]
  \centering
  \includegraphics[width=0.23\linewidth]{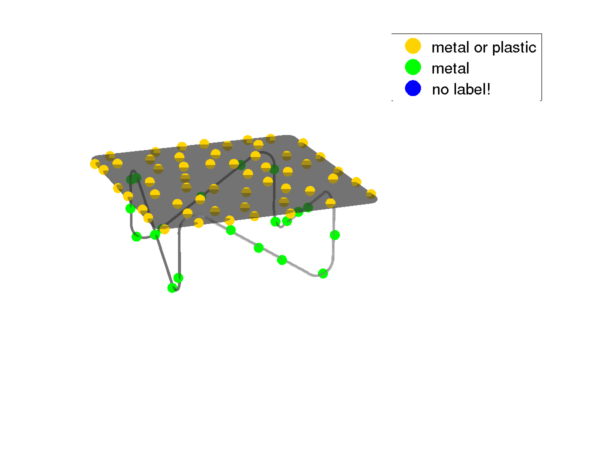}
  \includegraphics[width=0.23\linewidth]{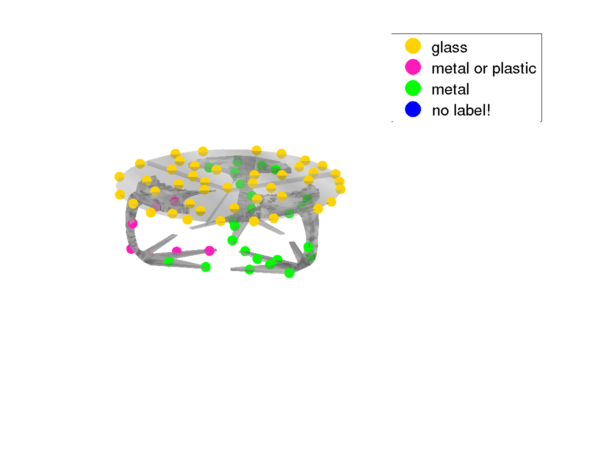}
  \includegraphics[width=0.23\linewidth]{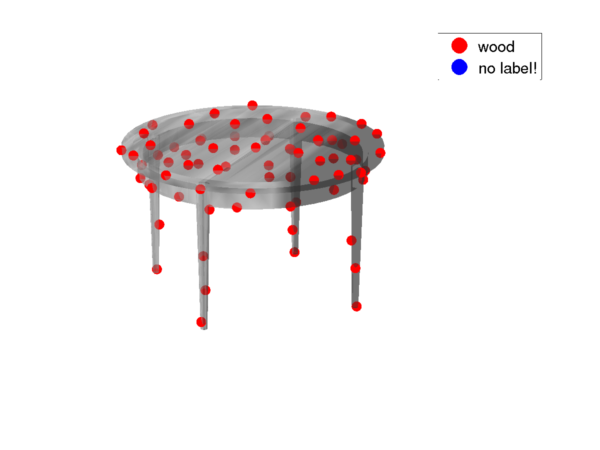}
  \includegraphics[width=0.23\linewidth]{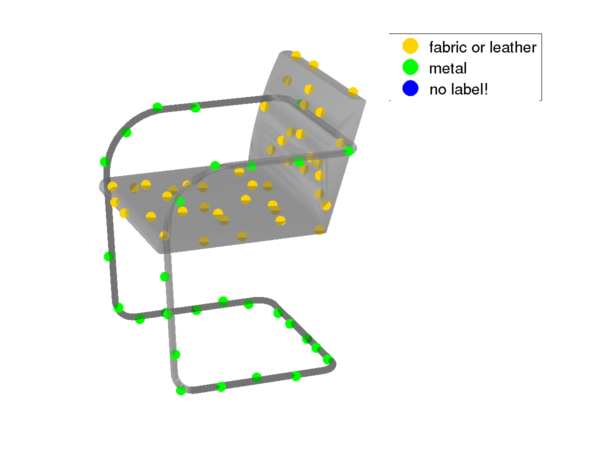}
  \includegraphics[width=0.23\linewidth]{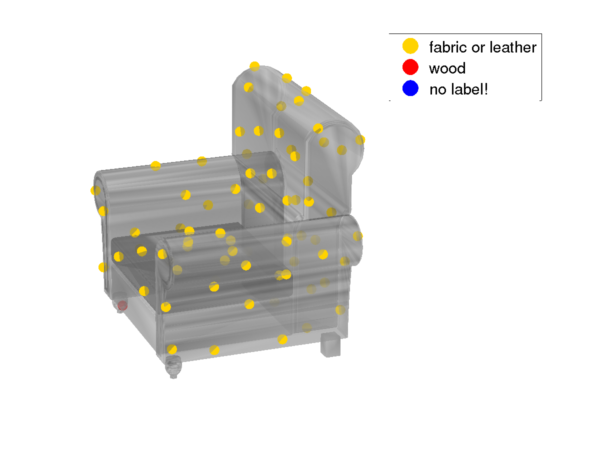}
  \includegraphics[width=0.23\linewidth]{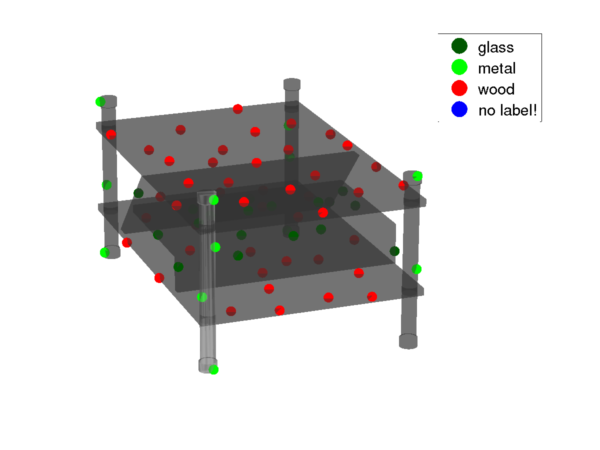}
  \includegraphics[width=0.23\linewidth]{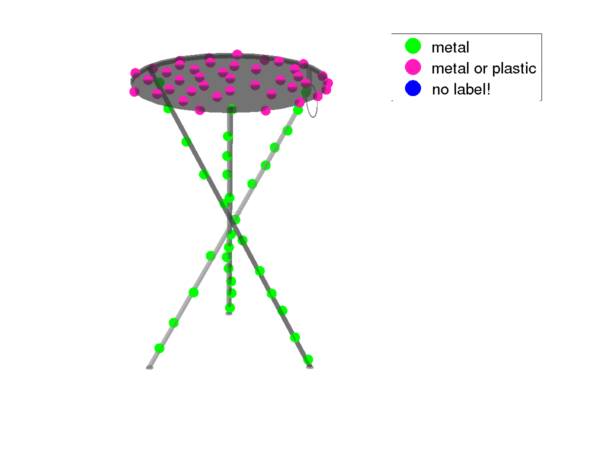}
  \includegraphics[width=0.23\linewidth]{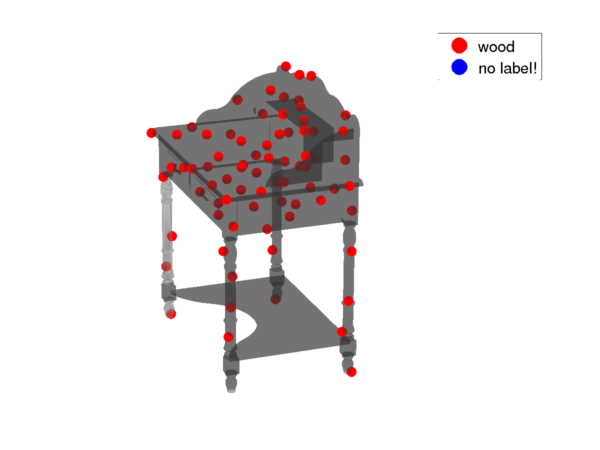}
  \includegraphics[width=0.23\linewidth]{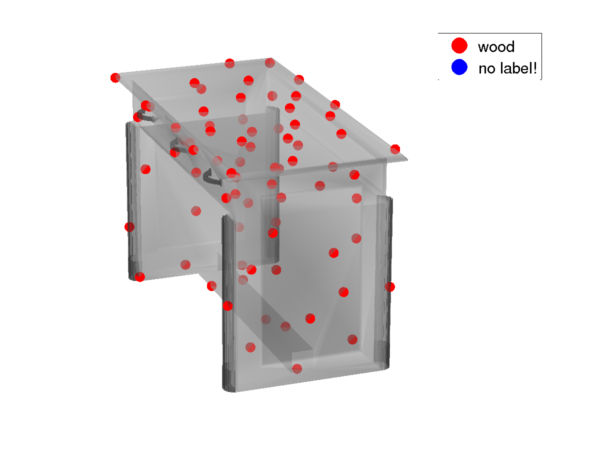}
  \includegraphics[width=0.23\linewidth]{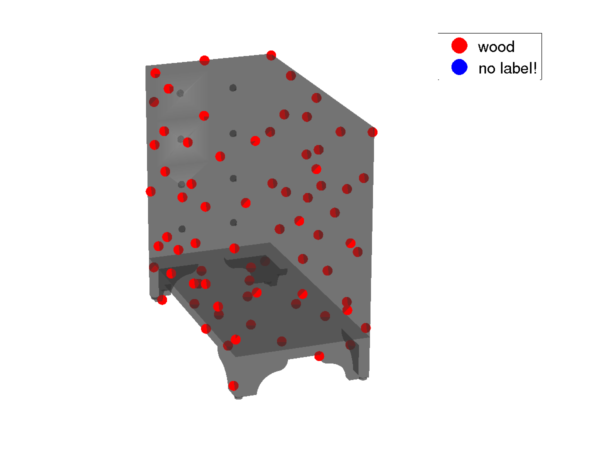}
  \includegraphics[width=0.23\linewidth]{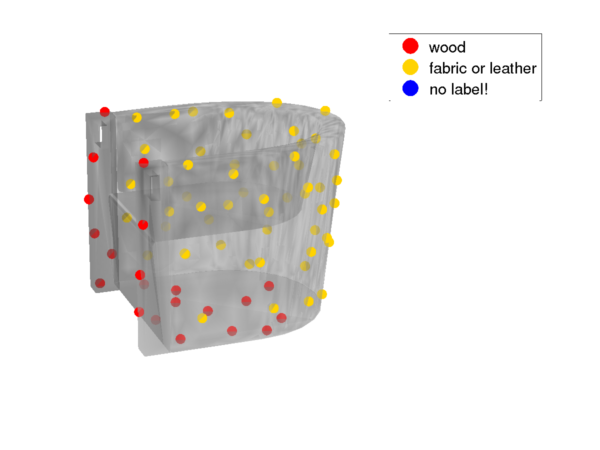}
  \includegraphics[width=0.23\linewidth]{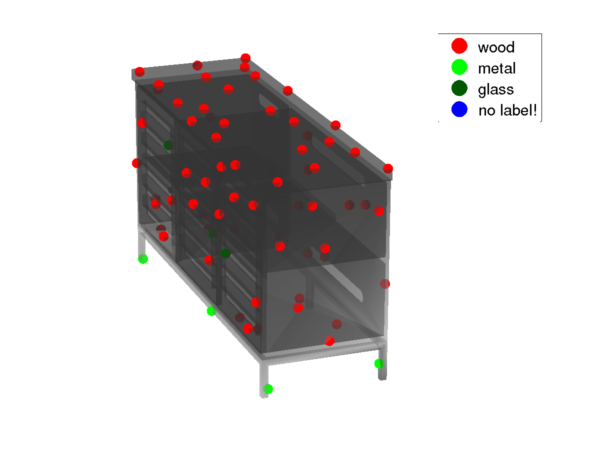}
  \includegraphics[width=0.23\linewidth]{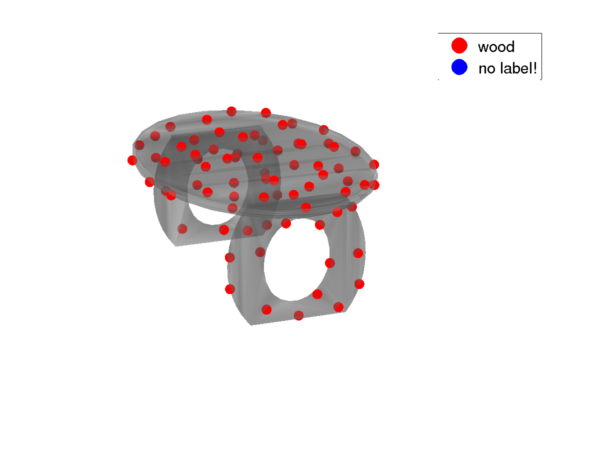}
  \includegraphics[width=0.23\linewidth]{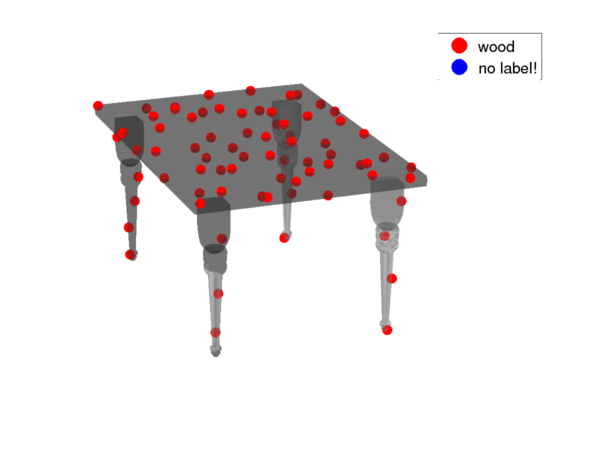}
  \includegraphics[width=0.23\linewidth]{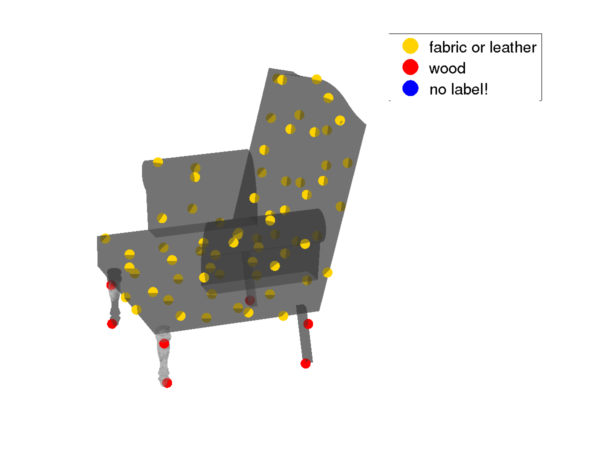}
  \includegraphics[width=0.23\linewidth]{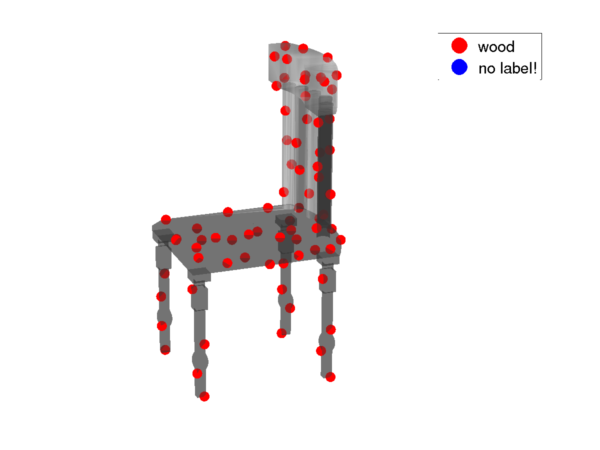}
  \includegraphics[width=0.23\linewidth]{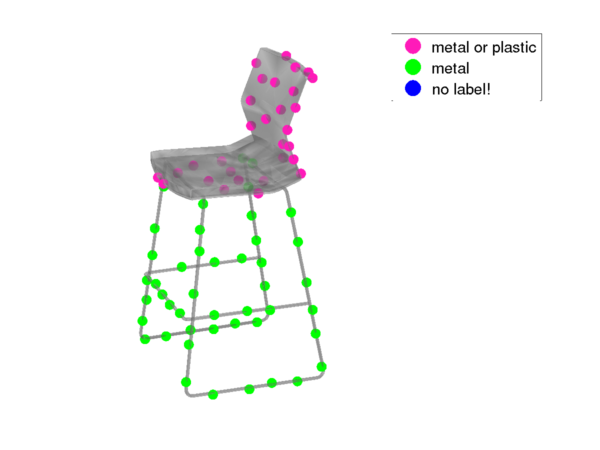}
  \includegraphics[width=0.23\linewidth]{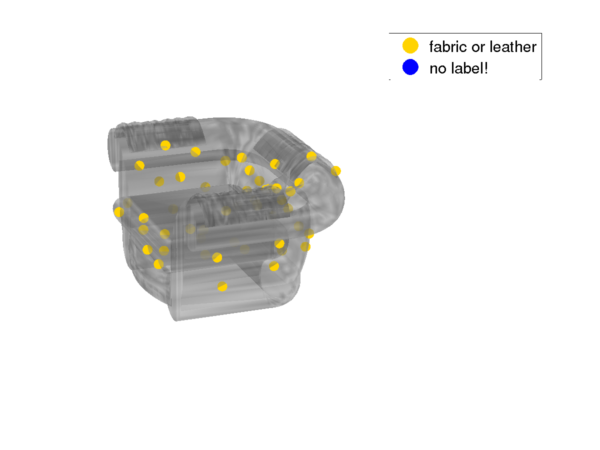}
  \includegraphics[width=0.23\linewidth]{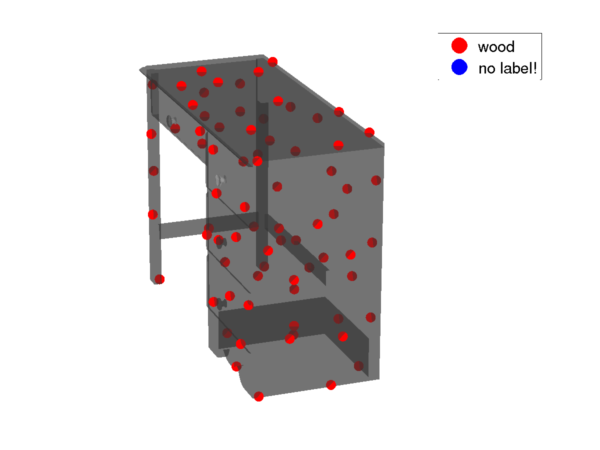}
  \includegraphics[width=0.23\linewidth]{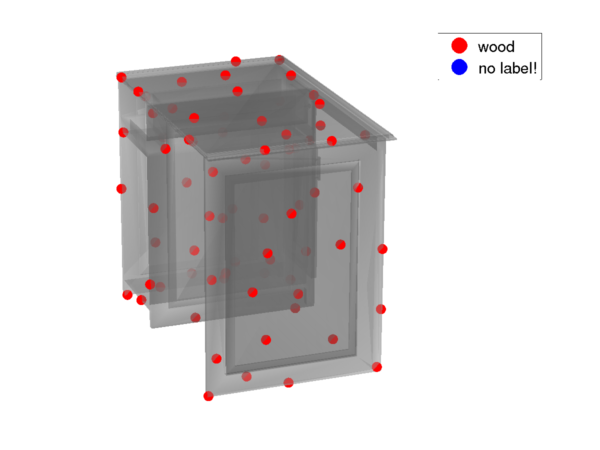}
  \includegraphics[width=0.23\linewidth]{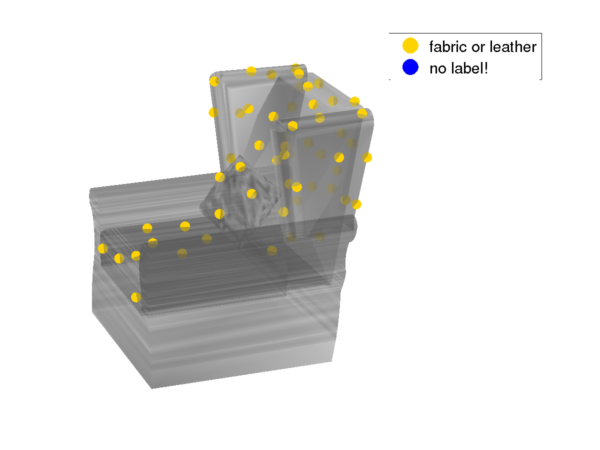}
  \includegraphics[width=0.23\linewidth]{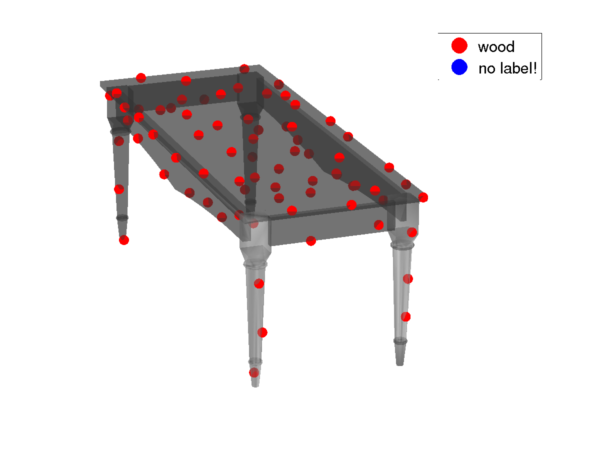}
  \includegraphics[width=0.23\linewidth]{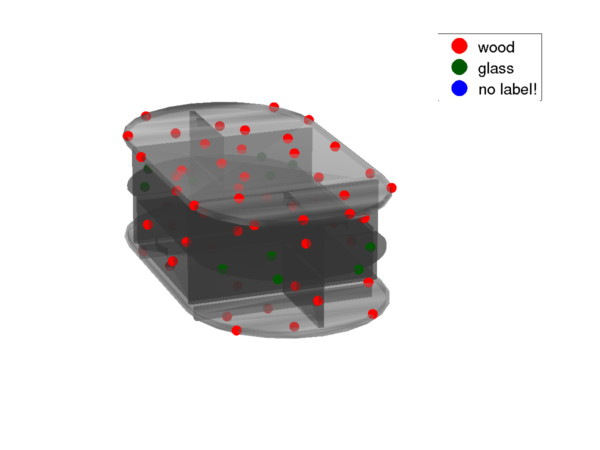}
  \includegraphics[width=0.23\linewidth]{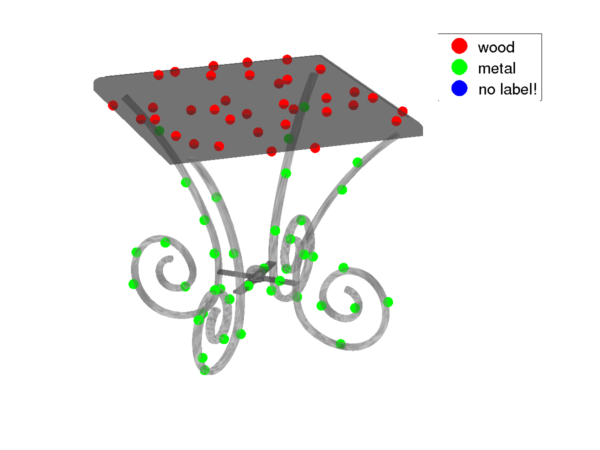}
  \includegraphics[width=0.23\linewidth]{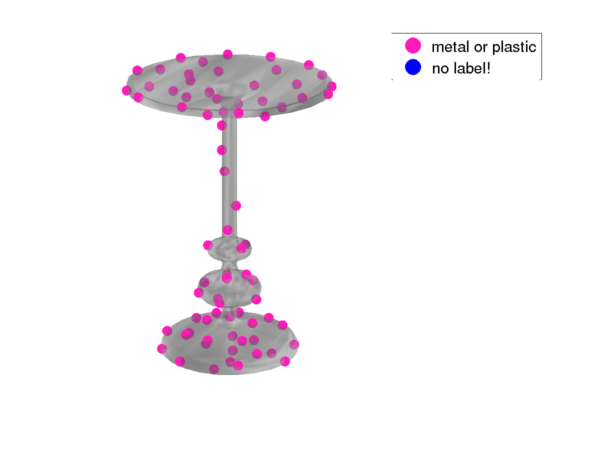}
  \includegraphics[width=0.23\linewidth]{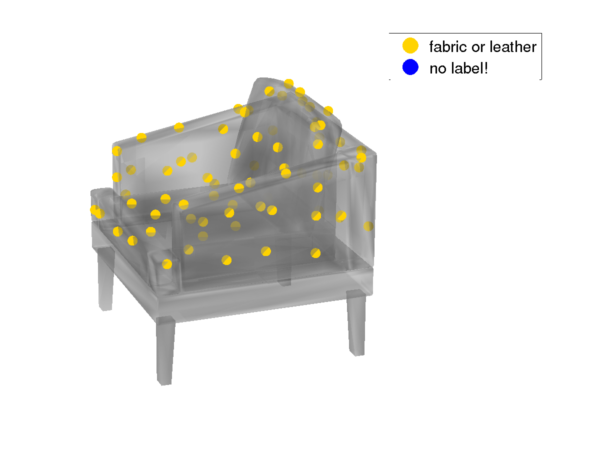}
  \includegraphics[width=0.23\linewidth]{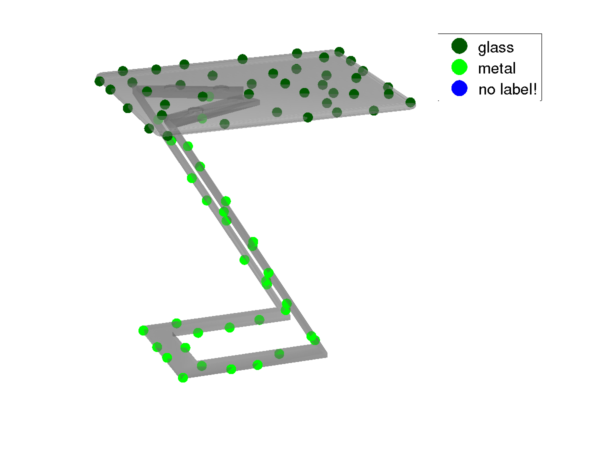}
  \includegraphics[width=0.23\linewidth]{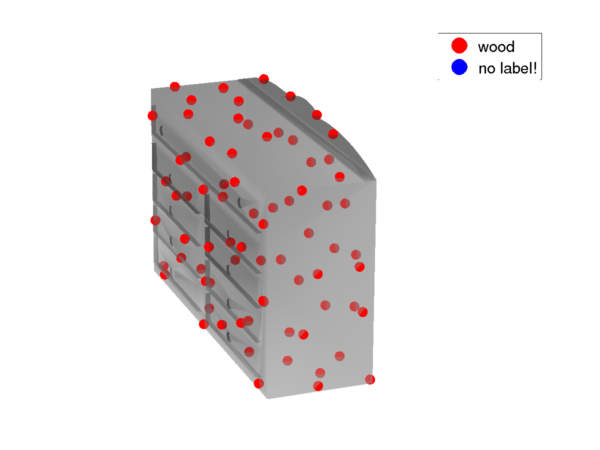}
  \vspace{-1mm} \caption{\small Small sample of crowdsourced dataset. 75
  training point samples with ground truth labels per shape are shown. Please
refer to legend for each shape for labels.} 
 \label{fig:mturkgt} 
 \vspace{-2mm}
\end{figure*}

Figure ~\ref{fig:benchmarkgt} shows a small sample of our benchmark test shapes
with ground truth labels. Figure ~\ref{fig:benchmarkpointpred} shows per-point
predictions for each of the 1024 test point samples on these benchmark test
shapes. Figure ~\ref{fig:benchmarkpartpred} shows per-part predictions after the
1024 point predictions are smoothed with our symmetry-aware CRF. Figure
~\ref{fig:mturkgt} shows a small sample of our MTurk crowdsourced data with the
75 training point samples per shape shown.

{\small
\bibliographystyle{ieee}
\bibliography{paper}
}

\end{document}